\documentclass[conference]{IEEEtran}
\usepackage{cite}
\usepackage{amsmath,amssymb,amsfonts}
\usepackage{subfigure}
\usepackage{bbm}
\usepackage{algorithmic}
\usepackage{graphicx}
\usepackage{textcomp}
\usepackage{xcolor}
\usepackage{tabularx}
\usepackage{hyperref}
\usepackage{multirow}
\def\BibTeX{{\rm B\kern-.05em{\sc i\kern-.025em b}\kern-.08em
    T\kern-.1667em\lower.7ex\hbox{E}\kern-.125emX}}

\begin{document}

\title{On Calibration of Graph Neural Networks\\ for Node Classification}

%\author{\IEEEauthorblockN{Anonymous Authors}}

\author{\IEEEauthorblockN{1\textsuperscript{st} Tong Liu}
\IEEEauthorblockA{\textit{LMU Munich} \\
Munich, Germany\\
tong.liu@physik.uni-muenchen.de}\\
\IEEEauthorblockN{3\textsuperscript{rd} Mitchell Joblin}
\IEEEauthorblockA{\textit{Siemens AG}\\
Munich, Germany \\
mitchell.joblin@siemens.com}
\and
\IEEEauthorblockN{1\textsuperscript{st} Yushan Liu}
\IEEEauthorblockA{\textit{Siemens AG, LMU Munich} \\
Munich, Germany \\
yushan.liu@siemens.com}\\
\IEEEauthorblockN{4\textsuperscript{th} Hang Li}
\IEEEauthorblockA{\textit{Siemens AG, LMU Munich} \\
Munich, Germany \\
hang.li@siemens.com}
\and
\IEEEauthorblockN{2\textsuperscript{nd} Marcel Hildebrandt}
\IEEEauthorblockA{\textit{Siemens AG}\\
Munich, Germany \\
marcel.hildebrandt@siemens.com}\\
\IEEEauthorblockN{5\textsuperscript{th} Volker Tresp}
\IEEEauthorblockA{\textit{Siemens AG, LMU Munich} \\
Munich, Germany \\
volker.tresp@siemens.com}
}

\maketitle

\begin{abstract}
Graphs can model real-world, complex systems by representing entities and their interactions in terms of nodes and edges. To better exploit the graph structure, graph neural networks have been developed, which learn entity and edge embeddings for tasks such as node classification and link prediction. These models achieve good performance with respect to accuracy, but the confidence scores associated with the predictions might not be calibrated. That means that the scores might not reflect the ground-truth probabilities of the predicted events, which would be especially important for safety-critical applications. Even though graph neural networks are used for a wide range of tasks, the calibration thereof has not been sufficiently explored yet. We investigate the calibration of graph neural networks for node classification, study the effect of existing post-processing calibration methods, and analyze the influence of model capacity, graph density, and a new loss function on calibration. Further, we propose a topology-aware calibration method that takes the neighboring nodes into account and yields improved calibration compared to baseline methods.
\end{abstract}


\section{Introduction}
Learning graph representations for relational data structures has been gaining increasing attention in the machine learning community~\cite{chami2021ml,hamilton2017representation}. A graph is able to model real-world, complex systems by representing entities as nodes and interactions between them as edges.
%Each edge stands for a true, known fact, e.\,g., the edge \textit{(Publication 1, cites, Publication 2)} indicates that \textit{Publication 1} contains a reference to \textit{Publication 2} in its text. If all edges in the graph are of the same type $\textit{cites}$, it is sufficient to denote the edge by \textit{(Publication 1, Publication 2)}. 
Since the information in graphs is often incomplete, e.g., missing node attributes or edges, relevant graph-related tasks for attaining new knowledge include node classification and link prediction. 
A variety of graph neural network (GNN) models have been developed~\cite{kipf2016semi, velivckovic2017graph, klicpera2019predict}, which learn node and edge embeddings in a low-dimensional vector space. Subsequently, these embeddings can be used to solve downstream tasks like node classification. 
Usually, the focus here lies on maximizing the accuracy -- the proportion of nodes that are classified correctly. 
GNNs achieve good performance with respect to accuracy but are also black boxes and lack interpretability. 

Most machine learning models output confidence scores associated with the predictions, and the concept of calibration captures the idea that the score should reflect the ground-truth probability of the prediction's correctness. For example, if 100 instances have a score of 0.6 for a specific class $k$, then 60 instances are expected to actually be of class $k$. 
A real-world application is autonomous driving, where the model should not only be aware that the object in front of the car is more likely to be a plastic bag than a pedestrian but also know how much more likely it is. A score distribution of 0.99 for plastic bag and 0.01 for pedestrian or 0.51 for plastic bag and 0.49 for pedestrian could have a huge influence on the next action of the car. 
Generally, calibrated scores lead to a better interpretation of the results and increase the trustworthiness of machine learning models, which is especially important in safety-critical domains.

The calibration of deep neural networks has been addressed in several works~\cite{guo2017calibration,thulasidasan2019mixup,tomani2021trustworthy,muller2019does}. The calibration of GNNs, however, has not been sufficiently explored yet, and existing calibration methods do not exploit the graph structure. Due to the different architectures of GNNs compared to neural networks, GNNs might exhibit different calibration characteristics.
In this work, we are interested in the following research questions:

\textbf{R1.}
How are GNNs calibrated for the node classification task, and are existing calibration methods sufficient to calibrate GNNs?

\textbf{R2.}
How do model capacity (width and depth) and graph density influence the calibration?

\textbf{R3.}
Can a calibration error term be added to the loss function in a straightforward way to improve the calibration without hurting the accuracy?

\textbf{R4.}
Can the incorporation of topological information improve calibration?

To better understand the calibration properties of GNNs, we conduct an empirical analysis of several GNN models in a node classification setting. 
Based on our experimental finding that the nodes in the graph express different levels of over- and underconfidence, we propose a topology-aware calibration method that takes the neighboring nodes into account.
Our contributions are summarized as follows:
\begin{itemize}
    \item We inspect the calibration of five representative GNN models on three benchmark citation datasets for node classification. 
    \item We analyze the influence of model capacity, graph density, and a new loss function on the calibration of GNNs. 
    %\item We apply five existing post-processing calibration methods, which are able to improve the calibration in most cases.
    \item We propose a calibration method that takes the graph topology into account and yields improved calibration compared to state-of-the-art post-processing calibration methods.
\end{itemize}
In Section~\ref{sec:background}, we define the necessary concepts and summarize related work. The existing GNNs and calibration methods used in this work are also described briefly. An experimental study on the calibration of GNNs is presented in Section~\ref{sec:experiments} ($\rightarrow$ \textbf{R1}, \textbf{R2}, \textbf{R3}). In Section~\ref{sec:ratio} ($\rightarrow$ \textbf{R1}, \textbf{R4}), we propose a topology-aware calibration method and show experimental results compared to state-of-the-art calibration baselines. The results are discussed in Section~\ref{sec:discussion}.

\section{Background}
\label{sec:background}

\subsection{Definitions}
\subsubsection{Node classification on graphs}
An undirected graph is defined as $G = (\mathcal{V} ,\mathcal{E})$, where $\mathcal{V}$ is the set of nodes and $\mathcal{E}$ the set of edges. An edge $e = \{i,j\} \in \mathcal{E}$ connects the two nodes $i$ and $j$ in the graph.
The information about the edges can be encoded in an adjacency matrix $\mathbf{A} \in \left \{0, 1\right \}^{\left|\mathcal{V}\right| \times \left|\mathcal{V}\right|}$. With $\mathbf{A}_{ij}$ being the entry in the $i$-th row and $j$-th column of $\mathbf{A}$, we define $\mathbf{A}_{ij} = 1$ if  $\{i,j\} \in \mathcal{E}$ and $\mathbf{A}_{ij} = 0$ otherwise\footnote{We identify nodes and indices to ease the notation.}. Moreover, we define $\mathcal{N}(i)$ as the set of neighbors of node $i$. For attributed graphs, where each node $i$ is associated with a $d$-dimensional feature vector $\mathbf{X}_i \in \mathbb{R}^d$, we denote the feature matrix by $\mathbf{X} \in \mathbb{R}^{|\mathcal{V}| \times d}$.

The goal of the node classification task is to assign each node $i \in \mathcal{V}$ a class label $\hat{y}_i \in \mathcal{K} := \{1, 2, \dots, K\}$, where $K$ stands for the total number of classes.

\subsubsection{Calibration}
Let $\mathbf{H}_i \in \mathbb{R}^h$ denote the node embedding and $y_i \in \mathcal{K}$ the ground-truth label of sample (or node) $i\in \mathcal{V}$.
Let $g: \mathbb{R}^h \rightarrow [0,1]^K$ be a function that takes $\mathbf{H}_i$ as input and outputs a probability vector $g(\mathbf{H}_i)$, where $g(\mathbf{H}_i)_k$ represents the $k$-th element. 
The predicted class label for sample $i$ is given by 
$\hat{y}_i =\mathop{\arg\max}_{k \in \mathcal{K}}g(\mathbf{H}_i)_k$, where $\hat{p}_i =\mathop{\max}_{k \in \mathcal{K}}g(\mathbf{H}_i)_k$ is called the corresponding confidence score for $\hat{y}_i$.
Perfect calibration is defined as
$
    \mathbb{P}(\hat{y}_i=y_i \mid \hat{p}_i=p)=p
$
for all $p \in [0,1]$ and any sample $i$~\cite{guo2017calibration}.

\begin{figure*}[t]
\centering 
\subfigure{
\includegraphics[width=0.235\textwidth]{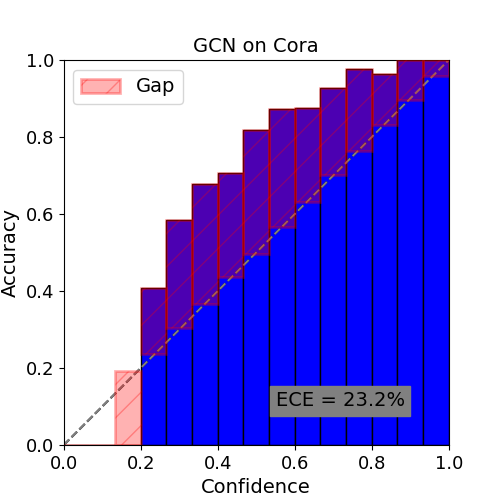}}
\subfigure{
\includegraphics[width=0.235\textwidth]{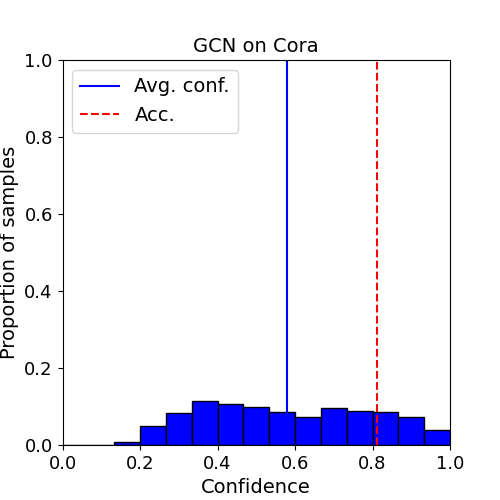}}
\subfigure{
\includegraphics[width=0.235\textwidth]{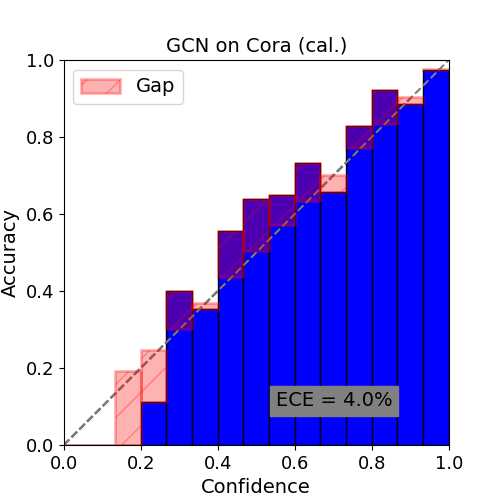}}
\subfigure{
\includegraphics[width=0.235\textwidth]{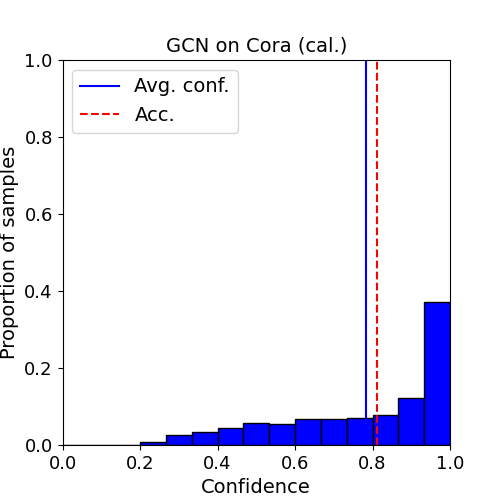}}
\caption{Reliability diagrams and corresponding confidence histograms for GCN on Cora. The two left plots show the results before calibration, while the two right plots show the results after calibration with temperature scaling. The diagonal line indicates perfect calibration.}
\label{fig:rds}
\end{figure*}

A reliability diagram~\cite{murphy1977reliability} plots accuracy against confidence to visualize the calibration of the model (see Fig. \ref{fig:rds}).
More formally, the samples are grouped into $M \in \mathbb{N}$ equally-spaced interval bins according to their confidences $\hat{p}_i$. For each bin $B_m$, $m \in \{1,2, \dots, M\}$, the accuracy and average confidence are calculated according to
\begin{align}
   \mathrm{acc}(B_m) &= \frac{1}{|B_m|}\sum_{i \in B_m}\mathbbm{1}[\hat{y}_i=y_i]
  \quad
   \mathrm{and} \\
  \quad
    \mathrm{conf}(B_m) &= \frac{1}{|B_m|}\sum_{i \in B_m}\hat{p}_i\;,
\end{align}
respectively, where $|B_m|$ denotes the number of samples in bin $B_m$ and $\mathbbm{1}$ the indicator function. In case of perfect calibration, the equation $\mathrm{acc}(B_m) = \mathrm{conf}(B_m)$ holds for all $m$.
Reliability diagrams also present a way to identify if the model is over- or underconfident. 
If the bars are above the diagonal line, it implies that the accuracy is higher than the average confidence, and the model is called underconfident. If the bars are below the diagonal, the model is overconfident.

The expected calibration error (ECE)~\cite{naeini2015obtaining} measures the miscalibration by averaging the gaps in the reliability diagram and is given by
\begin{equation}
\sum_{m=1}^{M}\frac{|B_m|}{N}|\mathrm{acc}(B_m)-\mathrm{conf}(B_m)|\;,
\label{eq:ece}
\end{equation}
where $N$ is the total number of samples.
%ECE is a widely used metric and convenient for comparing results with existing works, but it only takes the classes with the highest prediction probabilities into account. However, it might be of interest to measure the calibration with respect to all classes.

The marginal ECE (MECE) approximates the marginal calibration error~\cite{kumar2019verified}, which takes all classes into account. For each bin and every class $k$, it compares the average confidence of samples for class $k$ to the proportion of samples that has as ground-truth label class $k$. The MECE is defined as
\begin{equation}
\sum_{k=1}^K w_k \sum_{m=1}^{M} \frac{1}{N}\left|\sum_{i \in B_m}\mathbbm{1}[y_i=k] - \sum_{i \in B_m}g(\mathbf{H}_i)_k\right|\;,
\label{eq:mece}
\end{equation}
where $w_k$ is a class-dependent weight factor, which is set to $1/K$ if all classes are equally important.
%In case the bins are equally spaced, the MECE can become arbitrarily low for balanced datasets by rescaling all scores to be around $1/K$. To prevent this issue, we use an implementation in which all bins have an equal number of samples.

\subsection{Related work}
Guo et al.~\cite{guo2017calibration} showed that modern neural networks are miscalibrated and tend to be overconfident, i.\,e., the confidence scores are higher than the proportions of correct predictions. They proposed temperature scaling, a single-parameter variant of Platt scaling~\cite{platt1999probabilistic}, to calibrate the results. Several other methods were introduced to improve the calibration of deep neural networks (e.\,g., mixup training~\cite{thulasidasan2019mixup} and FALCON~\cite{tomani2021trustworthy}). 
Methods that improve calibration by preventing overconfidence include label smoothing~\cite{muller2019does} and focal loss~\cite{mukhoti2020calibrating, charoenphakdee2021focal}. 
% also show good results for overconfident models. 
% rather than underconfident GNNs in citation datasets. 
In GNNs, calibration issues have only been studied recently. 
A first evaluation of GNNs was done by Teixeira et al.~\cite{teixeira2019graph}, who performed experiments on multiple node classification datasets and concluded that GNNs are miscalibrated and existing calibration methods are not always able to improve the calibration to the desired extent.

\subsection{Methods}
\subsubsection{Graph neural networks}
Given an adjacency matrix $\mathbf{A}$ and a feature matrix $\mathbf{X}$, the idea of all GNNs is to learn node embeddings $\mathbf{H} \in \mathbb{R}^{|\mathcal{V}| \times h}$. The embedding for node $i$ is denoted by $\mathbf{H}_i \in \mathbb{R}^h$, which can be fed to a task-specific decoder $g$. For example, since we are concerned with node classification, we use a single-layer perceptron with softmax activation as decoder.
For our experiments, we select the widely used models graph convolutional network (GCN)~\cite{kipf2016semi}, graph attention network (GAT)~\cite{velivckovic2017graph}, and simple graph convolution (SGC)~\cite{wu2019simplifying}. Further, we consider graph filter neural network (gfNN)~\cite{nt2019revisiting}, a straightforward extension of SGC, and approximate personalized propagation of neural predictions (APPNP)~\cite{klicpera2019predict}, a model with state-of-the-art performance.

\textit{GCN} applies a normalized adjacency matrix with self-loops $\tilde{\mathbf{A}} = \hat{\mathbf{D}}^{-\frac{1}{2}} (\mathbf{A}+\mathbf{I}) \hat{\mathbf{D}}^{-\frac{1}{2}}$, where $\mathbf{I}$ is the identity matrix and $\hat{\mathbf{D}}$ the degree matrix of $\mathbf{A}+\mathbf{I}$. Concretely, the hidden layer of a GCN is formed according to
$
    \mathbf{H}^{(l+1)} 
    = \sigma(\tilde{\mathbf{A}}\mathbf{H}^{(l)}\mathbf{W}^{(l)}),
$
where ${\mathbf{W}^{(l)}}$ is a trainable weight matrix, $\sigma$ an activation function, and $\mathbf{H}^{(0)} := \mathbf{X}$. GCN aggregates information from a node's neighbors by computing the normalized sum of adjacent node embeddings.

\textit{GAT} differs from GCN in the neighbor aggregation function by introducing an attention mechanism that scales the importance of neighbors when summing over their embeddings. 
\iffalse
For each node $i$ and neighbor $j \in \mathcal{N}(i)$, the attention $\alpha_{ij}$ is computed as
{\fontsize{9}{10}
\begin{equation*}
    \alpha_{ij} = \frac{\exp(\mathrm{LeakyReLU}(\mathbf{a}^{T}[\mathbf{W}\mathbf{h}_{i};\mathbf{W}\mathbf{h}_{j}])}{\sum_{\tilde{j} \in \mathcal{N}(i)} \exp(\mathrm{LeakyReLU}(\mathbf{a}^{T}[\mathbf{W}\mathbf{h}_{i};\mathbf{W}\mathbf{h}_{\tilde{j}}]))}
    \label{2.9}
\end{equation*}}
where $\mathbf{a}$ is a trainable vector, $\mathbf{W}$ a trainable weight matrix, $\mathbf{h}_i$ the hidden embedding of node $i$, and $[\cdot\,;\cdot]$ the concatenation operation. 
\fi

\textit{SGC} is a GCN without nonlinear activation functions between the layers, resulting from the authors' conjecture that the good performance of GCNs comes from the aggregation of local neighborhood information and not from the application of nonlinear feature maps.

\textit{gfNN} extends SGC with a nonlinear layer $\sigma$ so that the node embeddings for layer $l$ are obtained from
$
\mathbf{H}^{(l)} = \sigma(\tilde{\mathbf{A}}^{l}\mathbf{X}\mathbf{W}),
$
where $\mathbf{W}$ is a trainable weight matrix.

\textit{APPNP} is based on the personalized PageRank (PPR) algorithm~\cite{page1998rank}. The node embeddings in layer $l+1$ are calculated via
$
\mathbf{H}^{(l+1)} = (1-\alpha)\tilde{\mathbf{A}}\mathbf{H}^{(l)} + \alpha \mathbf{H}^{(0)},
$
where $\alpha \in (0,1]$ is a hyperparameter and $\mathbf{H}^{(0)} := f(\mathbf{X})$, with $f$ being a trainable neural network.

\subsubsection{Calibration methods}
%Many existing approaches that were introduced for machine learning algorithms are post-processing methods that are tuned on the validation set. 
We consider the classical post-processing methods {histogram binning} \cite{zadrozny2001obtaining}, {isotonic regression}~\cite{zadrozny2002transforming}, and Bayesian binning into quantiles (BBQ)~\cite{naeini2015obtaining}, which is a refinement of histogram binning. Further, we include temperature scaling ~\cite{guo2017calibration} as a multiclass calibration method and {Meta-Cal}~\cite{ma2021meta}, a recently introduced approach with state-of-the-art performance.

\textit{Histogram binning} divides the confidence scores $\hat{p}_i$ into $M$ bins and assigns a new score $\hat{q}_m$ to each bin to represent the calibrated confidences. The scores $\hat{q}_m$ are learned by minimizing
$\sum_{m=1}^{M} \sum_{i \in B_m}(\hat{q}_m - y_i)^2$.
%which results in the optimal $\hat{q}_m$ being the accuracy of bin $B_m$.

\textit{Isotonic regression} learns a piecewise constant function $f$ 
%for obtaining the calibrated probabilities $\hat{q}_i = g(\hat{p}_i)$. The function $g$ is learned 
by minimizing $\sum_{m=1}^M \sum_{i \in B_m}(f(\hat{p}_i)-y_i)^2$. It is a generalization of histogram binning where the bin boundaries and scores are jointly optimized.

\textit{BBQ} extends histogram binning and learns a distribution $\mathbb{P}(\hat{q}_i \mid \hat{p}_i, \mathcal{D}_{\mathrm{val}})$ by marginalizing out all possible binnings, where $\mathcal{D}_{\mathrm{val}}$ is the validation set.

\textit{Temperature scaling} is a single-parameter extension of Platt scaling~\cite{platt1999probabilistic} for multiple classes. Given the output logit vector $\mathbf{z}$ before the softmax activation, a rescaling $\mathbf{z}/T$ depending on a temperature $T > 0$ is applied.
%to get the calibrated probabilities.
%$$\mathrm{softmax}\left(\frac{\mathbf{z}}{T}\right)=\frac{\exp(\frac{\mathbf{z}}{T})}{\sum_{k=1}^{K}\exp(\frac{\mathbf{z}_k}{T})}\;.$$
%The parameter $T$ is optimized with respect to NLL on the validation set. For $T>1$, it "softens" the confidence, making the model change from overconfident to a more calibrated state. 
%Since softmax is a monotonic function, temperature scaling does not affect the ranking of the classes' probabilities, and the accuracy stays unchanged.

\textit{Meta-Cal} combines temperature scaling as a base model with a bipartite ranking model to weaken the limitation of accuracy-preserving calibration methods. By investigating two practical constraints (miscoverage rate control and coverage accuracy control), the goal is to improve calibration depending on the bipartite ranking while controlling the accuracy.

\section{Experimental study}
\label{sec:experiments}
\subsection{Setup}
\noindent \textbf{Experiments} 
We first inspect the calibration of GNN models on benchmark citation datasets, where we take the best hyperparameter and training settings from the corresponding original papers. 

Then, we empirically analyze the influence of model capacity (width and depth) on calibration.
It has been observed that stacking too many GCN layers drastically worsens the performance, which is partly attributed to a phenomenon called oversmoothing~\cite{li2018deeper}. Oversmoothing happens when repeated neighbor aggregation leads to similar node embeddings in the graph, and various methods have been proposed to tackle this problem~\cite{klicpera2019predict,xu2018representation}. In the following, we investigate if increasing model depth also affects calibration.

One of the core mechanisms of GNNs is the message aggregation from neighboring nodes.
%, where the number of edges for the node plays a role. 
We examine how graph density, i.\,e., the ratio of the number of edges in the graph to the number of maximum possible edges, influences the calibration performance. 

Finally, we also test a new loss function \eqref{eq:loss}
%Since the cross-entropy loss does not automatically optimize for calibration, we try a new loss function
that combines the standard cross-entropy loss $L_{\mathrm{ce}}$ with an ECE-inspired term $L_{\mathrm{cal}}$ for optimizing the calibration. We define $L_{\mathrm{cal}}$ as the cross entropy between the confidence of the sample and the accuracy of its corresponding bin, where the idea is that the confidence should stay close to the accuracy.
Given the original cross-entropy loss $L_{\mathrm{ce}}$, we define the new loss as
\begin{table}[t]
\setlength{\tabcolsep}{0.5em}
\setlength{\extrarowheight}{0.1em}
\caption{Dataset statistics.}
\centering
%\resizebox{0.35\textwidth}{!}{
\begin{tabular}{c|cccccc}
\hline
Dataset & $K$ & $d$ & $|\mathcal{V}|$ & $|\mathcal{E}|$ &  Label rate \\
\hline
Cora     & 7 & 1,433 & 2,708  & 5,429  & 0.052  \\
Citeseer & 6 & 3,703 & 3,327  & 4,732 & 0.036 \\
Pubmed   & 3 & 500  & 19,717 & 44,338  & 0.003 \\
\hline
\end{tabular}
\label{tab:statistics}
\end{table}
\begin{align}
\label{eq:loss}
L &= \alpha L_{\mathrm{ce}} + (1- \alpha) L_{\mathrm{cal}}\quad \mathrm{with}\\
L_{\mathrm{cal}} &= 
- \sum_{i=1}^{N}\mathrm{acc}(B_{m}(i))\cdot \log(\hat{p}_i)\;,
\end{align}
where $\alpha \in (0,1)$ and $B_{m}(i)$ denotes the bin that sample $i$ belongs to.

For the experiments on width, depth, graph density, and the new loss function, we focus on GCN and GAT, two of the basic and most widely used GNN models.

\noindent \textbf{Datasets}
Cora, Citeseer, and Pubmed\footnote{\url{https://github.com/kimiyoung/planetoid}} are three commonly used benchmark datasets for node classification. 
They are citation networks, where nodes represent scientific publications and edges between pairs of nodes correspond to one publication citing the other. Each node comes with a $d$-dimensional feature vector that indicates the presence of words from a predefined vocabulary. The class label of a node is the topic of the corresponding publication. Similar to previous works~\cite{kipf2016semi,velivckovic2017graph}, we operate under a semi-supervised setting, where only a small amount of labeled data is available during training. The statistics of the datasets are summarized in Table~\ref{tab:statistics}.

\noindent \textbf{Implementation}
The GNN models are implemented using the PyTorch-Geometric library\footnote{\url{https://github.com/pyg-team/pytorch_geometric}}.
The bin number for calculating the ECE and MECE is set to 15. 
More information about the hyperparameters and experimental settings can be found in the supplementary material\footnote{Source code and supplementary material available at \url{https://github.com/liu-yushan/calGNN}.}.

\subsection{Results}
\label{sec:gnn_calibration}

\noindent \textbf{Uncalibrated results} 
We run all GNNs on the three citation datasets and show the uncalibrated performance with respect to accuracy, ECE, and MECE in Table~\ref{tab:uncal_performance}.
The method APPNP is best on Cora and Pubmed in terms of accuracy (second-best on Citeseer), and it is also best calibrated on the datasets Citeseer and Pubmed. For Cora, gfNN has the lowest ECE and MECE. 
All models except for gfNN\footnote{The original paper trains for 50 epochs without early stopping. A different training setting might stabilize the results more.} have stable calibration values with small standard deviations. The worst method with respect to the calibration performance is SGC, which is, apart from the softmax activation for normalization, the only linear model. Adding a nonlinear layer as in gfNN results in better calibration.
Moreover, we find that GAT outperforms GCN in terms of ECE and MECE in two of three datasets.
%ECE and MECE seem to be correlated so that MECE increases with increasing ECE, while the magnitude of MECE is overall smaller.
\begin{table}[t]
\setlength{\tabcolsep}{0.5em}
\setlength{\extrarowheight}{0.1em}
\caption{Uncalibrated performance with respect to accuracy, ECE, and MECE (mean±SD over 100 independent runs). The best results are displayed in bold. }
\label{tab:uncal_performance}
\centering
%\resizebox{0.5\textwidth}{!}{
\begin{tabular}{cc|ccc}
\hline
Dataset  & Model                      & Acc.             & ECE       & MECE  \\ \hline
\multirow{5}{*}{Cora}  
& GCN   & 81.43±0.60           & 23.51±1.89        &7.01±0.46  \\
& GAT   & 83.14±0.39            & 17.26±1.09          & 5.15±0.30 \\
& SGC & 81.19±0.05 &26.03±0.16&	7.78±0.08 \\
& gfNN  & 78.73±5.04          & \textbf{6.45±2.44}        & \textbf{3.16±1.45}          \\
& APPNP & \textbf{83.68±0.36}   & 14.90±0.69      & 4.73±0.17             \\
\hline
\multirow{5}{*}{Citeseer}   
& GCN   &  71.32±0.70         & 21.80±1.21       & 8.57±0.32 \\
& GAT   & 70.99±0.60           & 18.92±1.05        & 7.66±0.30     \\ 
& SGC   & \textbf{72.46±0.15}&	53.59±0.14&	19.15±0.00\\
& gfNN  & 67.33±6.58         &  15.50±4.47       & 8.40±1.26          \\
& APPNP & 72.10±0.38 & \textbf{11.93±0.80}          & \textbf{5.37±0.28}              \\
\hline
\multirow{5}{*}{Pubmed}  
& GCN   & 79.23±0.43        & 10.62±1.28       & 7.29±0.84  \\
& GAT   & 79.05±0.38       & 14.37±0.48  & 9.89±0.23 \\
& SGC   & 78.72±0.04 &	22.40±0.04 &	14.98±0.02 \\
& gfNN  & 77.94±2.32  & 6.04±2.90&  5.23±2.65 \\
& APPNP & \textbf{80.09±0.25}   & \textbf{4.38±0.74}         & \textbf{3.59±0.41}            \\
\hline       
\end{tabular}
\end{table}
\begin{figure}[t]
\centering 
\subfigure{
\includegraphics[width=0.23\textwidth]{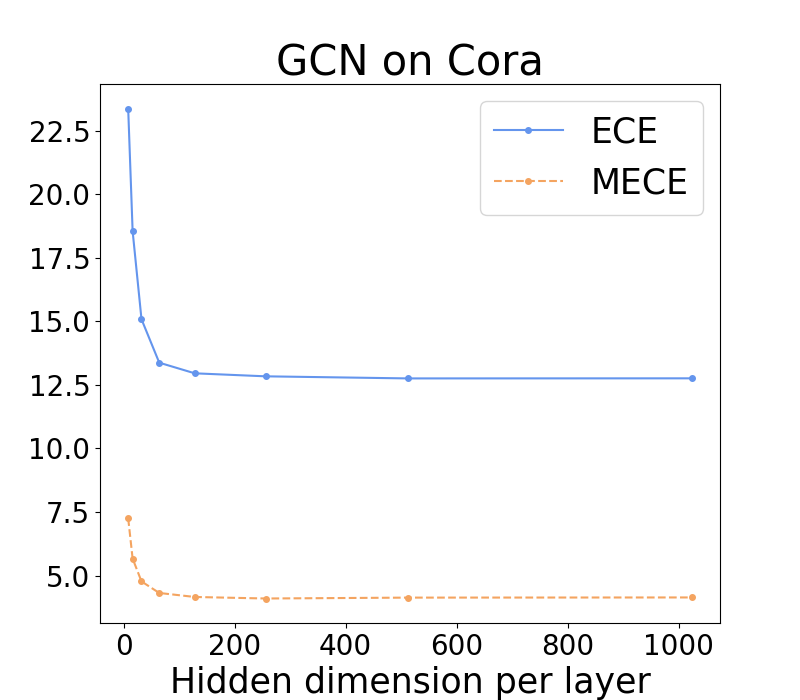}}
\hspace{-5mm}
\subfigure{
\includegraphics[width=0.23\textwidth]{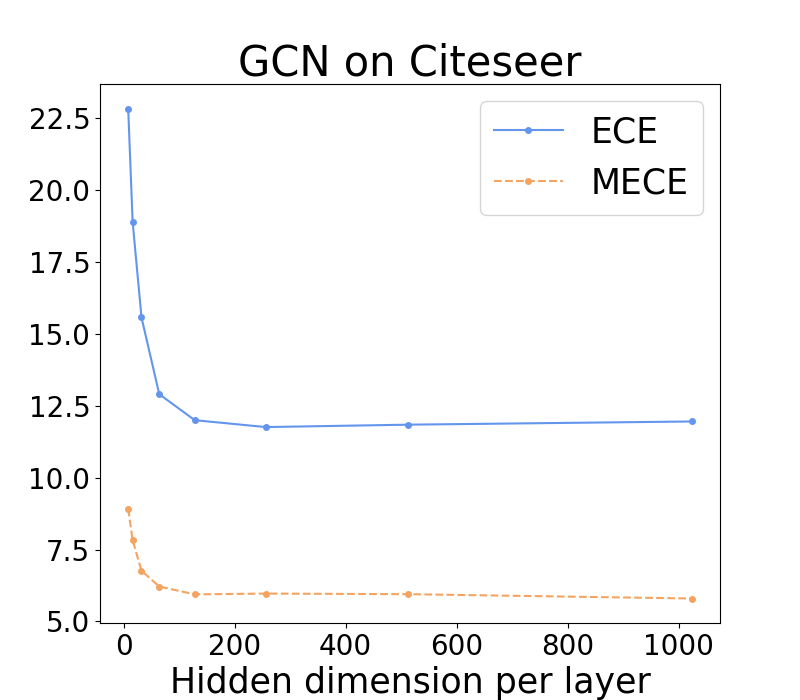}}
\subfigure{
\includegraphics[width=0.23\textwidth]{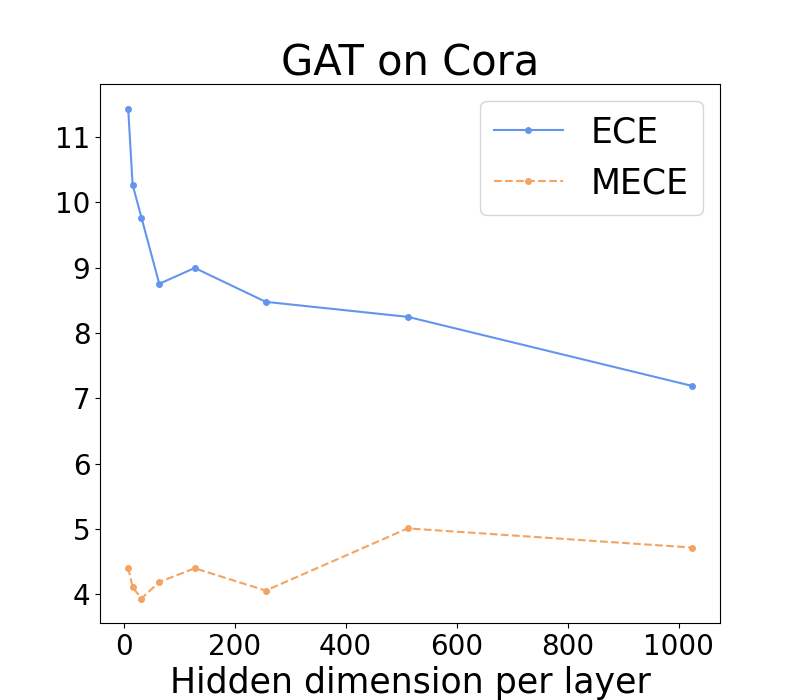}}
\hspace{-5mm}
\subfigure{
\includegraphics[width=0.23\textwidth]{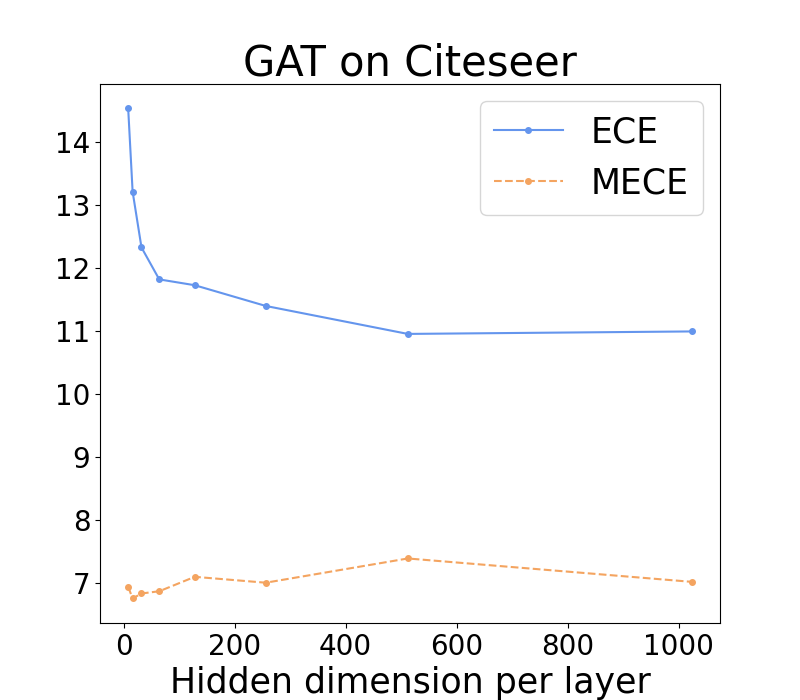}}
\caption{Varying model width (hidden dimension per layer).}
\label{fig:width}
\end{figure}
\begin{figure}[t]
\centering 
\subfigure{
\includegraphics[width=0.23\textwidth]{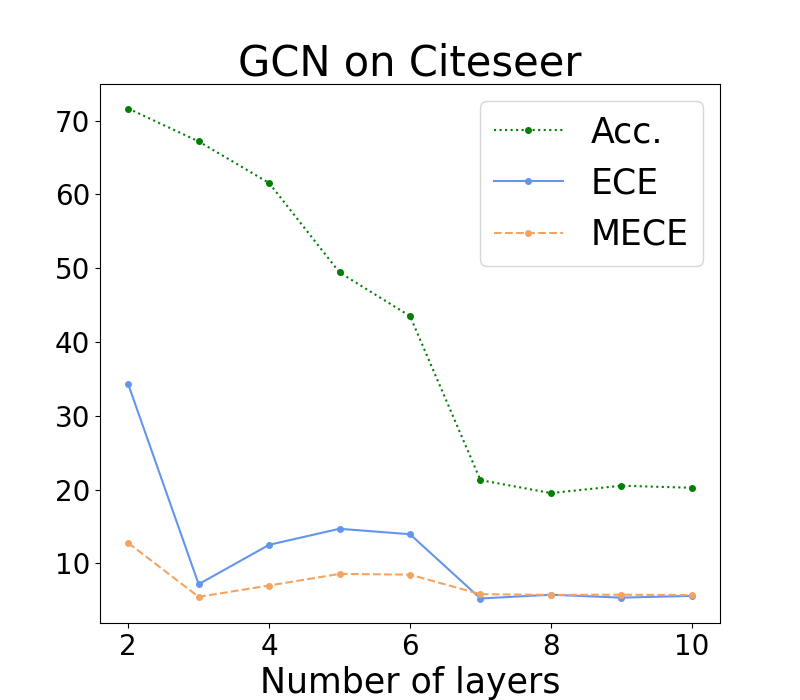}}
\hspace{-5mm}
\subfigure{
\includegraphics[width=0.23\textwidth]{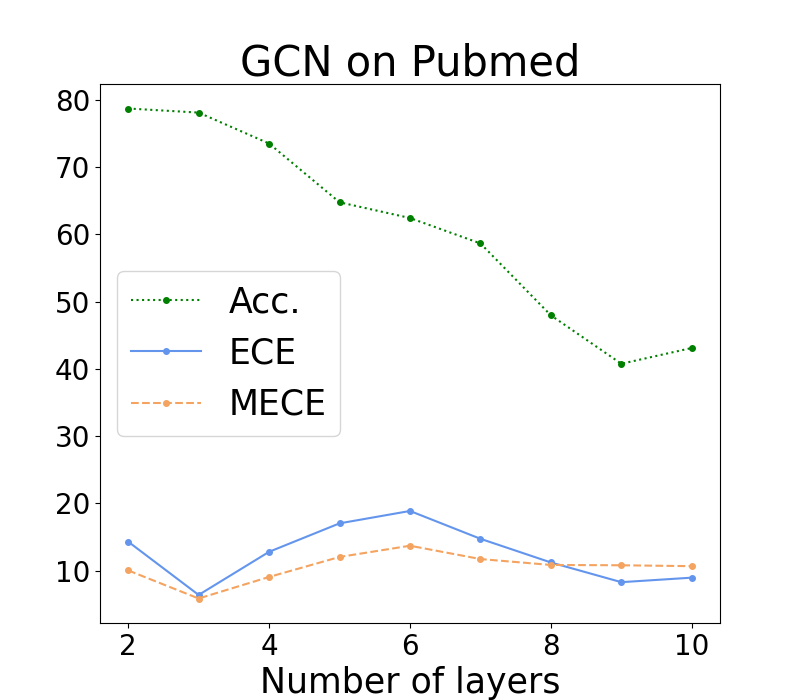}}
\subfigure{
\includegraphics[width=0.23\textwidth]{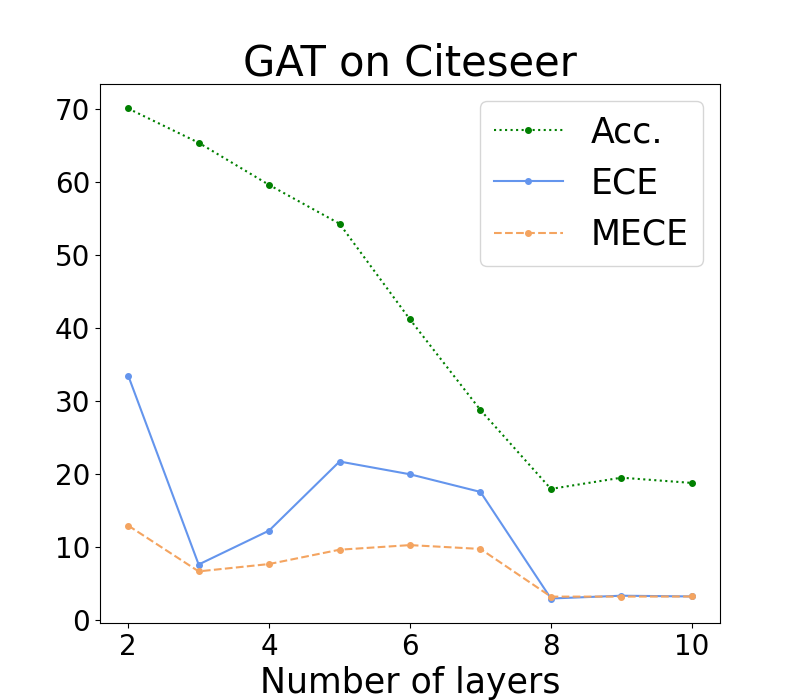}}
\hspace{-5mm}
\subfigure{
\includegraphics[width=0.23\textwidth]{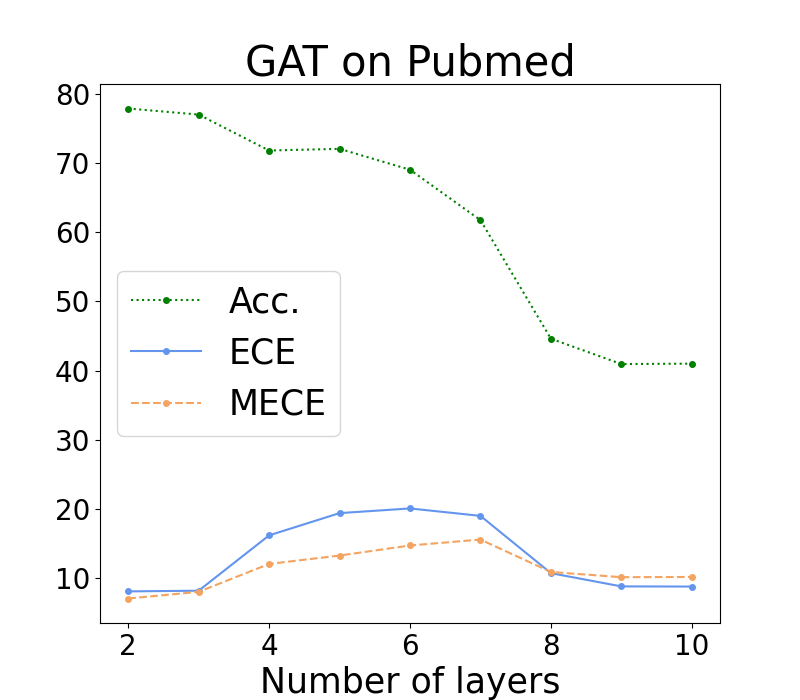}}
\caption{Varying depth (number of layers).}
\label{fig:depth}
\end{figure}

\noindent \textbf{Influence of width} We compare the calibration of GCN and GAT for varying model width, i.\,e., the number of hidden dimensions per layer.
%, which is given by $\{2^{i} \mid 3\leq i \leq 10\}$. 
While the accuracy basically stays constant, the ECE and MECE
decrease with increasing number of hidden dimensions initially (see Fig.~\ref{fig:width}). When a certain width is reached, the calibration values stagnate or slightly increase again. Generally, wider networks tend to be better calibrated.
\begin{table*}[t]
\setlength{\tabcolsep}{0.5em}
\setlength{\extrarowheight}{0.2em}
\caption{Uncalibrated performance of GCN and GAT under the standard and the new loss function (mean±SD over 10 independent runs). The better results when comparing the two loss functions are underlined.}
\label{tab:cal_loss}
\centering
%\renewcommand{\arraystretch}{1.2}
%\resizebox{0.6\textwidth}{!}{
\begin{tabular}{cc|ccc|ccc}
\hline
Dataset  & Model                      & Acc. ($L_{\mathrm{ce}}$)                    & ECE ($L_{\mathrm{ce}}$)          & MECE ($L_{\mathrm{ce}}$) & Acc. ($L$) & ECE ($L$) & MECE ($L$)  \\ \hline
\multirow{2}{*}{Cora}  
& GCN   & 81.43±0.60           & 23.51±1.89        &7.01±0.46   & \underline{81.81±0.85} & \underline{14.91±2.7} & \underline{4.64±0.56}           \\
& GAT   &  \underline{83.14±0.39}            & 17.26±1.09          & 5.15±0.30       & 82.73±0.40 &\underline{5.29±1.31} & \underline{2.41±0.32}      \\
\hline
\multirow{2}{*}{Citeseer}   
& GCN   &  71.32±0.70         & 21.80±1.21       & 8.57±0.32          &\underline{71.67±0.50} & \underline{14.65±0.42} & \underline{6.43±0.26} \\
& GAT   &  70.99±0.60           & 18.92±1.05        & 7.66±0.30         &\underline{71.13±0.44} &\underline{9.39±0.97} &\underline{4.58±0.40}       \\ 
\hline
\multirow{2}{*}{Pubmed}  
& GCN   & \underline{79.23±0.43}        & 10.62±1.28       & 7.29±0.84    & 79.00±0.37 &\underline{7.50±0.91} &\underline{5.60±0.73}    \\
& GAT   & \underline{79.05±0.38}       & 14.37±0.48  & 9.89±0.23& 78.88±0.40 & \underline{9.58±0.79} & \underline{6.91±0.63}\\
\hline       
\end{tabular}
\end{table*}

\noindent \textbf{Influence of depth} We investigate the influence of model depth, i.\,e., the number of layers, on the calibration performance (see Fig. \ref{fig:depth}). Oversmoothing becomes particularly pronounced when the test accuracy decreases significantly with increasing number of layers.
The ECE first improves when changing from two to three layers, then it increases again until five or six layers. Using an even larger model depth, the ECE eventually decreases again.
%The best results for both classification and calibration are obtained with 2 or 3 layers. 
%This is in accordance with most practical settings, where the number of GNN layers is often set to 2 for node classification tasks on various datasets.

\begin{figure}[t]
\centering 
\hspace{-8mm}
\subfigure{
\includegraphics[width=0.25\textwidth]{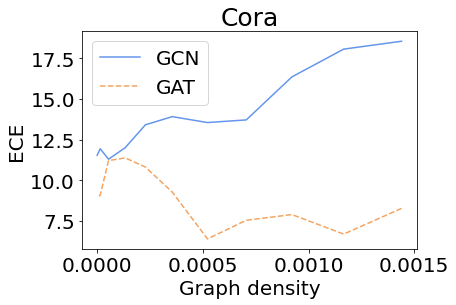}}\hspace{-6mm}
\subfigure{
\includegraphics[width=0.22\textwidth]{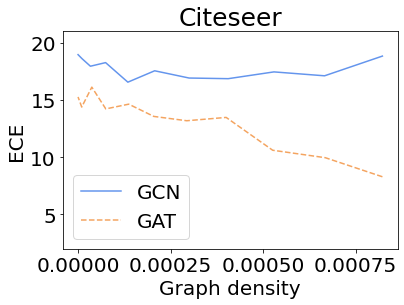}}\hspace{-13mm}
\includegraphics[width=0.22\textwidth]{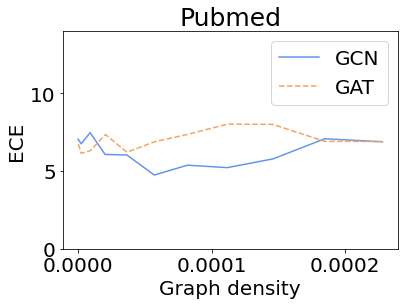}
\caption{Influence of graph density. The graph density is the ratio of the number of edges in the graph to the number of maximum possible edges.}
\label{fig:density}
\end{figure}

\noindent \textbf{Influence of graph density}
% We want to know how graph density, i.\,e., the ratio of the number of edges in the graph with respect to the number of maximum possible edges, influences the calibration performance. 
For this experiment, we remove different proportions of edges randomly from the dataset, ranging from $0\%$ (original dataset) to $100\%$ (no graph structure at all). The models GCN and GAT only differ in the aggregation mechanism, i.\,e., GAT introduces attention coefficients to weight the importance of neighbors.  
%To also identify the influence of attention, we take the same hyperparameters for both models 
%and set the number of attention heads for GAT to 8.
%The average node degree ranges from $0$, which is equivalent to a multilayer perceptron, to their original values. 
The results are shown in Fig.~\ref{fig:density}. Similar to Table~\ref{tab:uncal_performance}, the ECE of GAT is consistently lower than the ECE of GCN on Cora and Citeseer, while on Pubmed, GCN expresses partly better calibration. Generally, the graph density of Pubmed is the lowest. It might be that the attention weights in GAT are beneficial for calibration and especially useful when enough edges exist in the graph.

\noindent \textbf{Influence of new loss function}
Table~\ref{tab:cal_loss} compares the results of the standard cross-entropy loss $L_{\mathrm{ce}}$ and the new loss function $L$ from \eqref{eq:loss}, which contains a calibration error term. The new loss $L$ improves the model calibration in all cases while keeping the accuracy at the same level or even slightly increasing the accuracy. 

\noindent \textbf{Underconfidence vs. overconfidence}
Taking the best hyperparameter and training settings from their corresponding publications, all GNNs exhibit underconfidence on all three datasets, i.\,e., the confidence scores are lower than the accuracy of the predictions (see Fig.~\ref{fig:rds} and figures in the supplementary material).
In some cases, however, we find that the model changes from underconfidence to overconfidence if it is trained without early stopping. 
\begin{figure*}[t]
\centering 
\subfigure{
\includegraphics[width=0.32\textwidth]{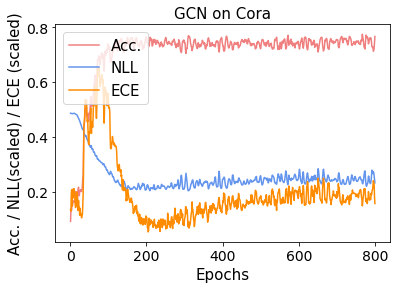}}\hspace{3mm}
\subfigure{
\includegraphics[width=0.32\textwidth]{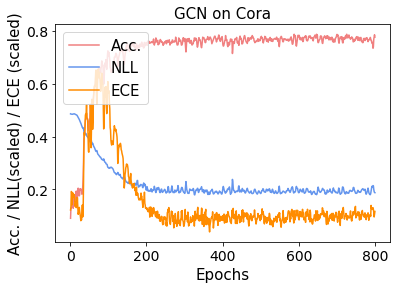}}
\caption{Test accuracy, scaled test NLL, and scaled test ECE for GCN on Cora, with a weight decay of $5\text{e-}4$ (left) and $7.5\text{e-}4$ (right).}
\label{fig:under_over_conf_epochs}
\end{figure*}
\begin{figure*}[t]
\centering 
\subfigure{
\includegraphics[width=0.23\textwidth]{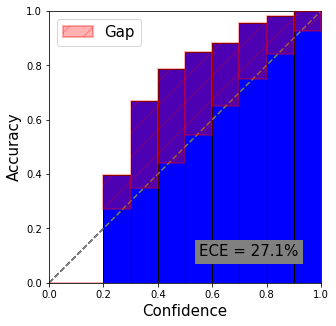}}
\subfigure{
\includegraphics[width=0.23\textwidth]{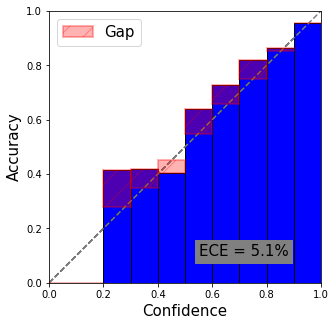}}
\subfigure{
\includegraphics[width=0.23\textwidth]{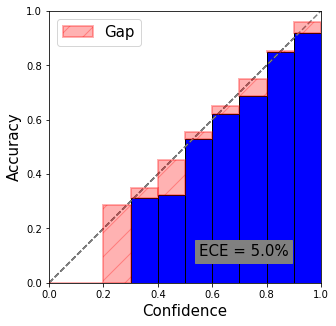}}
\subfigure{
\includegraphics[width=0.23\textwidth]{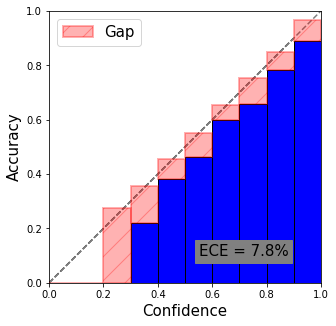}}
\caption{Reliability diagrams for GCN on Cora during the training process. From left to right, the corresponding number of epochs is 100, 200, 300, and $400$. }
\label{fig:under_over_conf_rd}
\end{figure*}
%We set the learning rate to $0.01$. 

The left plot in Fig.~\ref{fig:under_over_conf_epochs} shows the test accuracy, scaled test negative log-likelihood (NLL), and scaled test ECE for a $4$-layer GCN on Cora during training, with a weight decay set to $5\text{e-}4$. 
Around epoch $150$, the NLL and accuracy become stable, while the ECE is still improving. At this point, GCN is underconfident, as shown in the left-most reliability diagram in Fig.~\ref{fig:under_over_conf_rd}.
In the epochs between $200$ and $300$, the ECE gains the best performance when the model changes from underconfidence to overconfidence (see the two diagrams in the middle of Fig.~\ref{fig:under_over_conf_rd}). After epoch $300$, GCN starts to overfit with repect to the ECE, while the NLL and accuracy remain rather unchanged. 
During this process, overconfidence aggravates, and the ECE increases to $7.8\%$ in epoch 400, which is displayed in the right-most diagram in Fig. \ref{fig:under_over_conf_rd}. 

In summary, GCN first optimizes NLL and accuracy during training, then fits the confidence scores, and eventually starts to overfit regarding the ECE without influencing the NLL and accuracy.
When we slightly increase the weight decay to $7.5\text{e-}4$ (see right plot in Fig.~\ref{fig:under_over_conf_epochs}), the ECE stabilizes after reaching the optimal value, and the values of NLL and accuracy also stay in a smaller range compared to the left plot.

\section{Ratio-binned scaling for calibrating GNNs}
\label{sec:ratio}

\subsection{Same-class-neighbor ratio}
From our experiments, we find that GNN models tend to be underconfident. Even though the overall model exhibits underconfidence, there might be differences depending on node-level properties, which have not been considered before. Especially for graph data, the topology could provide structural information that are useful for calibration. For node classification, the class labels and properties of a node's neighbors have a significant influence on the classification.
We calculate for each node $i$ the same-class-neighbor ratio, i.\,e., the proportion of neighbors that have the same class as node $i$, and develop a new binning scheme that groups samples into bins based on the same-class-neighbor ratio for calibration.

\begin{figure*}[t]
\centering 
\subfigure{
\includegraphics[width=0.22\textwidth]{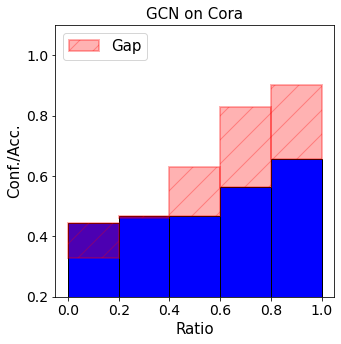}}
\subfigure{
\includegraphics[width=0.22\textwidth]{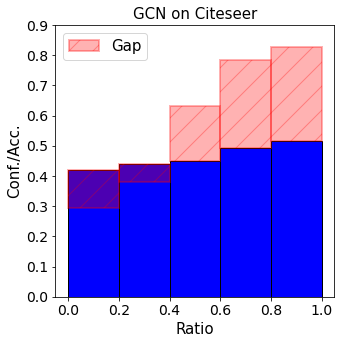}}
\subfigure{
\includegraphics[width=0.22\textwidth]{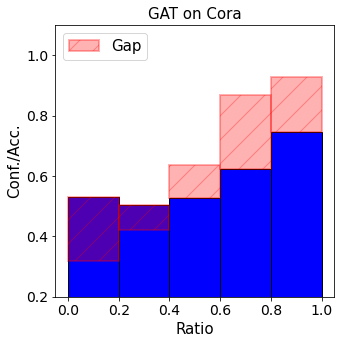}}
\subfigure{
\includegraphics[width=0.22\textwidth]{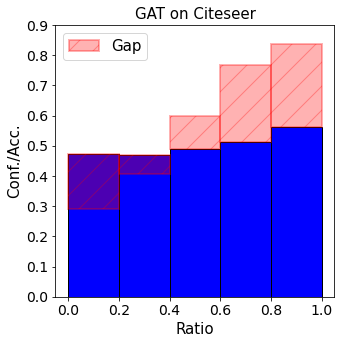}}
\caption{The nodes are grouped according to their same-class-neighbor ratios. The blue bar represents the average confidence and the gap the difference between average confidence and accuracy in each bin.}
\label{fig:ratio}
\end{figure*}

To evaluate the correlation between the same-class-neighbor ratio and the confidence of a model, we calculate the ratio for each node based on the ground-truth labels. In Fig.~\ref{fig:ratio}, we group the nodes into $5$ equally-spaced interval bins according to their ratios. 
Employing a trained GNN model, we compute the output of the classifier $g$ for each node and draw the average confidence of the samples in each bin as a blue bar. The gap illustrates the difference between the average confidence and the accuracy in each bin.
We observe that the average confidence increases with the same-class-neighbor ratio, where bins with higher ratios express underconfidence and bins with lower ratios overconfidence. 
Consequently, a binning scheme that groups samples depending on their same-class-neighbor ratios would take the graph structure into account and allow for an adaptive calibration depending on the confidence level of each bin.

\subsection{Ratio-binned scaling}
We propose ratio-binned scaling (RBS), a topology-aware method, which first approximates the same-class-neighbor ratio for each sample, then groups the samples into $M$ bins, and finally learns a temperature for each bin to rescale the confidence scores. 

In the semi-supervised setting, we only know the labels of a small number of nodes and therefore cannot use the true labels for binning. One natural option is to replace the nodes' ground-truth labels with their confidence scores for estimating the same-class-neighbor ratio. More precisely, we define the estimated ratio for node $i$ as 
\begin{equation}
\hat{r}(i) = \frac{1}{\left | \mathcal{N}(i) \right |} \sum_{j \in \mathcal{N}(i)} g(\mathbf{H}_j)_{\hat{y}_i} \quad \in [0,1]\;,
\end{equation}
where $\mathbf{H}_j$ is the node embedding of node $j$, which is learned by a GNN model, and $g$ is the classifier. $g(\mathbf{H}_j)_{\hat{y}_i}$ denotes the confidence score of node $j$ corresponding to the class $\hat{y}_i$ that is predicted for the central node $i$.

\begin{table*}[t]
\setlength{\tabcolsep}{0.5em}
\setlength{\extrarowheight}{0.1em}
\caption{Calibrated performance with respect to ECE (mean±SD over 100 independent runs). The best GNN model for each calibration method is underlined. The best calibration method for each GNN model is displayed in bold.}
\label{tab:cal_performance}
\centering
%\resizebox{0.6\textwidth}{!}{
\begin{tabular}{cc|c|cccccc|c}
\hline
Dataset  & Model    & Uncal. & His. bin.  & Iso. reg.  & BBQ      & Tem. scal.   & Meta-Cal & RBS & RRBS\\\hline
\multirow{5}{*}{Cora}   
& GCN  &23.51±1.89 & 4.50±0.76   & 3.94±0.66  &4.53±0.64  & \textbf{3.82±0.60}
  &4.09±0.65  & 3.90±0.61 & 3.10±0.63\\
& GAT                        & 17.26±1.09 & 4.86±0.62         & 4.04±0.59         & 4.18±0.60 & 3.53±0.66
  & \underline {\textbf{3.28±0.65}} &  3.34±0.63 & 2.67±0.54\\
& SGC &26.03±0.16 & 4.38±0.30         & 4.21±0.42 & 4.35±0.21 &4.05±0.11
  & 4.02±0.35          &  \textbf{3.55±0.07} & 2.57±0.11\\
& gfNN                      &  6.45±2.44 & \underline {3.80±0.86}         & \textbf{3.72±0.78} & 4.15±1.22& 3.74±1.34
  &  4.07±1.52      &  3.77±0.96 & 3.39±1.22\\
& APPNP                      & 14.90±0.69& 4.20±0.62       & \underline{3.43±0.60}&  \underline{3.90±0.57} &  \underline{3.14±0.50}
  &  3.48±0.57    &  \underline{\textbf{3.01±0.53}} &2.68±0.44\\
\hline
\multirow{5}{*}{Citeseer}  
& GCN                        & 21.80±1.21 & \underline{4.61±0.82} & \underline{\textbf{4.46±0.93}}       &  5.30±1.09  & 4.86±0.76
  & 5.04±0.90     &  4.99±0.75 & 4.11±0.88 \\
& GAT                        & 18.92±1.05 &  5.00±0.73         & 4.90±0.69 & \underline{5.04±0.71} & 5.92±0.58
  & 6.08±0.61         & \textbf{4.45±0.73} & 4.71±0.67\\
& SGC                        & 53.59±0.14 & 7.55±0.23      & 6.93±0.19 & 7.43±0.13 & \underline {4.47±0.19}
 & \underline {4.17±0.29}   & \textbf{\underline {4.04±0.17}} & 2.97±0.20    \\
& gfNN                       & 15.50±4.47 & \textbf{4.74±1.00} & 4.92±1.03        & 5.06±1.37 &  5.43±1.27
  & 5.45±1.32     &5.19±1.31 & 4.34±1.27\\
& APPNP                      & 11.93±0.80& 4.75±0.85          & \textbf{4.50±0.67} & 5.10±1.02 &  4.98±0.67
 &  5.29±0.67        &  5.08±0.69& 3.97±0.58\\
\hline
\multirow{5}{*}{Pubmed}   
& GCN                        & 10.62±1.28 &  4.69±0.78       & 4.76±0.77     & \underline{4.69±0.74} & 4.27±0.61
    & 4.99±1.11     &  \textbf{4.16±0.60}& 3.28±0.89 \\
& GAT                        & 14.37±0.48 & 4.85±0.89       & 4.93±0.78       & 5.70±1.03 & \underline {3.94±0.67}
  & 4.45±0.74  & \underline{\textbf{3.61±0.75}} &2.56±0.56\\
& SGC                        &22.40±0.04 & \underline {4.40±0.29}           & \underline{4.29±0.21}          & \underline {5.04±0.22}  & 4.13±0.12
  & 4.64±0.58    &\textbf{4.07±0.17} & 3.01±0.12   \\
& gfNN                       & 6.04±2.90 & 4.98±1.06        & 5.00±0.83         & 5.15±1.26  & 4.97±1.67
  & 6.06±1.95      & \textbf{4.91±1.65} & 3.78±1.03\\
& APPNP                      & 4.38±0.74 & 4.86±0.75         & 4.72±0.57           &  4.79±0.84 & 3.98±0.59
   & \underline{4.34±0.72}    & \textbf{3.80±0.60}  &2.60±0.47\\
\hline       
\end{tabular}
\end{table*}
For the ratio-based binning scheme, let $\{B_m \mid 1 \leq m \leq M\}$ be a set of bins that partitions the interval $[0,1]$ uniformly. After calculating the output for all nodes, each node $i$ is assigned to a bin according to its estimated same-class-neighbor ratio $\hat{r}(i)$, i.\,e., $B_1 = \{i \in \mathcal{V} \mid \hat{r}(i) \in [0, \frac{1}{M}]\}$ and $B_m = \{i \in \mathcal{V} \mid \hat{r}(i) \in (\frac{m-1}{M},\frac{m}{M}]\}$ for $m \in \{2, \dots, M\}$. 

Let the output of $g$ be in the form $g(\mathbf{H}_i) = \sigma(\mathbf{Z}_i) \in \mathbb{R}^K$ for node $i$, where $\sigma$ is the softmax function and $\mathbf{Z}_i$ the logits before normalization.  
For each bin $B_m$, $m \in \{1, \dots, M\}$, a temperature $T_m > 0$ is learned on the validation dataset. 

The calibrated confidence for a test node $j$ is then given by 
\begin{equation}
\mathbf{\hat{q}}_j = \sigma (\mathbf{Z}_{j}/T_{m}) \in [0,1]^K\quad \mathrm{if}\; j \in B_m.    
\end{equation}

We apply temperature scaling for calibrating the nodes in each bin, but it would also be possible to apply other post-processing calibration methods for obtaining calibrated scores. 

\subsection{Results}
Table~\ref{tab:cal_performance} summarizes the calibration performance of all considered post-processing calibration methods and our proposed method RBS in terms of ECE. All methods can improve the calibration of GNN models on Cora and Citeseer. In particular, the obtained ECE for a specific dataset and calibration method is rather similar for all GNNs regardless of the uncalibrated ECE. 
On Pubmed, most methods have difficulties improving calibration of APPNP, which already has low ECE. 
RBS gains the best performance in the majority of the cases and outperforms classical temperature scaling in $11$ out of $15$ experiments. 

Next to a good calibration performance, accuracy preservation is desirable for calibration methods. RBS and temperature scaling do not change the ranking of the classes and thus the accuracy stays unchanged. Meta-Cal trades good calibration for lower accuracy, while the other methods yield comparable or even slightly improved accuracy in some cases (see supplementary material).
%Compared to the results under the calibration loss in Table~\ref{table3.1}, only GAT ($L$) directly yields better calibration than the post-processing methods.

\subsection{Effectiveness of real-ratio-binned scaling}
Table~\ref{tab:cal_performance} also shows the calibrated results of real-ratio-binned scaling (RRBS), where we assume that the ground-truth labels are available for all nodes. RRBS outperforms the best calibration method in 14 out of 15 experiments. Although the correct labels are not accessible in the semi-supervised setting, the results still indicate the effectiveness of the intuition of our proposed method. 
%, which implies that RBS can lead to good calibration performance if the estimated ratio is close to the real same-class-neighbor ratio.

\section{Discussion}
\label{sec:discussion}

In general, the calibration performance depends on the specific GNN model and dataset, where all models perform best on Pubmed (see Table~\ref{tab:cal_performance}).
When using the hyperparameter and training settings from the original publications, all GNNs tend to be underconfident on all three datasets, in contrast to the finding that deep neural network models rather exhibit overconfidence~\cite{guo2017calibration,thulasidasan2019mixup}. However, when plotting the reliability diagrams for a varying number of epochs, we observe that in some cases, underconfidence changes to overconfidence when the number of epochs increases. It seems that underconfidence or overconfidence is not necessarily a property of the model architecture but is also dependent on the training setting.

%GATs are equivalent to transformers \cite{vaswani2017attention} when the input window of tokens given to the transformer is represented as a fully-connected graph. Alternatively, GAT can be seen as a transformer with an additional edge mask, which ensures that information is only propagated between nodes that are connected in the graph.
%, and as a downside, increasing complexity from  $O(|\mathcal{V}|^{2})$ to $O(|\mathcal{V}||\mathcal{\mathcal{E}}|)$, since generally the number of edges $|\mathcal{E}|$ is larger than the number of nodes $|\mathcal{V}|$ in a graph.
%Desai and Durrett~\cite{desai2020calibration} show that pre-trained transformers are well-calibrated in in-domain settings.
%Similarly, our results suggest that GAT is better calibrated than GCN, which might be more obvious in denser graphs. 

Most GNNs suffer from oversmoothing, which becomes apparent when increasing the number of layers in the model~\cite{klicpera2019predict,li2018deeper,xu2018representation}. We observe that for large numbers of layers, the accuracy drops significantly, while the ECE improves. Oversmoothing results in similar node embeddings, which might be uninformative for the model. In this case, the model would most likely learn the distribution of classes in the training data as confidence scores. Therefore, all samples would be grouped into one bin, resulting in low ECE if the test distribution is close to the training distribution. However, such a model does not make use of the underlying graph structure and is probably not useful for application.

The results for RBS and RRBS show the potential of a calibration method that takes the graph structure into account, where the binning scheme is constructed depending on node-level properties. RRBS almost always outperforms RBS, which suggests that RBS might be especially helpful for cases where the estimated ratios are close to the real ratios, i.e., for models with relatively high accuracy. The number of bins for RBS was chosen from $\{2,3,4\}$, and it seems that even a small number of bins can lead to improved calibration compared to classical temperature scaling.
It would further be interesting to apply RBS to other kinds of datasets, e.g., heterophilic graphs, where nodes from different classes are likely to be connected.

\section{Conclusion}
We investigated the calibration of graph neural networks for node classification on three benchmark datasets. Graph neural networks seem to be miscalibrated, where the exact calibration depends on both the dataset and the model. Existing post-processing calibration methods are able to alleviate the miscalibration but do not consider the graph structure. Based on our experimental finding that the nodes in the graph express different levels of over- or underconfidence depending on their same-class-neighbor ratios, we proposed the topology-aware calibration metohd ratio-binned scaling. It takes the predictions of neighboring nodes into account and shows better performance compared to state-of-the-art baselines.
%The analysis on the influence of model capacity and graph density yields that wider models and oversmoothing tend to hurt calibration, while attention-based GNNs might benefit from larger graph density. A combined loss function that also optimizes calibration can also decrease the expected calibration error but cannot directly outperform post-processing methods.
For future work, it would be interesting to gain a more theoretical understanding of the calibration properties and conduct experiments on larger and a wider variety of datasets.

\section*{Acknowledgment}
This work has been supported by the German Federal Ministry for Economic Affairs and Climate Action (BMWK) as part of the project RAKI under grant number 01MD19012C.

\bibliographystyle{IEEEtran}
\bibliography{mybibliography}

\begin{thebibliography}{10}
\providecommand{\url}[1]{#1}
\csname url@samestyle\endcsname
\providecommand{\newblock}{\relax}
\providecommand{\bibinfo}[2]{#2}
\providecommand{\BIBentrySTDinterwordspacing}{\spaceskip=0pt\relax}
\providecommand{\BIBentryALTinterwordstretchfactor}{4}
\providecommand{\BIBentryALTinterwordspacing}{\spaceskip=\fontdimen2\font plus
\BIBentryALTinterwordstretchfactor\fontdimen3\font minus
  \fontdimen4\font\relax}
\providecommand{\BIBforeignlanguage}[2]{{%
\expandafter\ifx\csname l@#1\endcsname\relax
\typeout{** WARNING: IEEEtran.bst: No hyphenation pattern has been}%
\typeout{** loaded for the language `#1'. Using the pattern for}%
\typeout{** the default language instead.}%
\else
\language=\csname l@#1\endcsname
\fi
#2}}
\providecommand{\BIBdecl}{\relax}
\BIBdecl

\bibitem{chami2021ml}
I.~Chami, S.~Abu-El-Haija, B.~Perozzi, C.~R\'{e}, and K.~Murphy, ``Machine
  learning on graphs: A model and comprehensive taxonomy,''
  \emph{arXiv:2005.03675}, 2021.

\bibitem{hamilton2017representation}
W.~L. Hamilton, R.~Ying, and J.~Leskovec, ``Representation learning on graphs:
  Methods and applications,'' \emph{Bulletin of the IEEE Computer Society
  Technical Committee on Data Engineering}, vol.~40, pp. 52--74, 2017.

\bibitem{kipf2016semi}
T.~N. Kipf and M.~Welling, ``Semi-supervised classification with graph
  convolutional networks,'' in \emph{Proceedings of the Fifth International
  Conference on Learning Representations}, 2017.

\bibitem{velivckovic2017graph}
P.~Veli{\v{c}}kovi{\'c}, G.~Cucurull, A.~Casanova, A.~Romero, P.~Li\`{o}, and
  Y.~Bengio, ``Graph attention networks,'' in \emph{Proceedings of the Sixth
  International Conference on Learning Representations}, 2018.

\bibitem{klicpera2019predict}
J.~Klicpera, A.~Bojchevski, and S.~Günnemann, ``Predict then propagate: Graph
  neural networks meet personalized {PageRank},'' in \emph{Proceedings of the
  Seventh International Conference on Learning Representations}, 2019.

\bibitem{guo2017calibration}
C.~Guo, G.~Pleiss, Y.~Sun, and K.~Q. Weinberger, ``On calibration of modern
  neural networks,'' in \emph{Proceedings of the Thirty-Fourth International
  Conference on Machine Learning}, 2017.

\bibitem{thulasidasan2019mixup}
S.~Thulasidasan, G.~Chennupati, J.~Bilmes, T.~Bhattacharya, and S.~Michalak,
  ``On mixup training: Improved calibration and predictive uncertainty for deep
  neural networks,'' in \emph{Proceedings of the Thirty-Third Conference on
  Neural Information Processing Systems}, 2019.

\bibitem{tomani2021trustworthy}
C.~Tomani and F.~Buettner, ``Towards trustworthy predictions from deep neural
  networks with fast adversarial calibration,'' in \emph{Proceedings of the
  Thirty-Fifth AAAI Conference on Artificial Intelligence}, 2021.

\bibitem{muller2019does}
R.~M{\"u}ller, S.~Kornblith, and G.~Hinton, ``When does label smoothing help?''
  in \emph{Proceedings of the Thirty-Third Conference on Neural Information
  Processing Systems}, 2019.

\bibitem{murphy1977reliability}
A.~H. Murphy and R.~L. Winkler, ``Reliability of subjective probability
  forecasts of precipitation and temperature,'' \emph{Applied Statistics},
  vol.~26, pp. 41--47, 1977.

\bibitem{naeini2015obtaining}
M.~P. Naeini, G.~F. Cooper, and M.~Hauskrecht, ``Obtaining well calibrated
  probabilities using {B}ayesian binning,'' in \emph{Proceedings of the
  Twenty-Ninth AAAI Conference on Artificial Intelligence}, 2015.

\bibitem{kumar2019verified}
A.~Kumar, P.~Liang, and T.~Ma, ``Verified uncertainty calibration,'' in
  \emph{Proceedings of the Thirty-Third Conference on Neural Information
  Processing Systems}, 2019.

\bibitem{platt1999probabilistic}
J.~C. Platt, ``Probabilistic outputs for support vector machines and
  comparisons to regularized likelihood methods,'' \emph{Advances in Large
  Margin Classifiers}, 1999.

\bibitem{mukhoti2020calibrating}
J.~Mukhoti, V.~Kulharia, A.~Sanyal, S.~Golodetz, P.~H.~S. Torr, and P.~K.
  Dokania, ``Calibrating deep neural networks using focal loss,'' in
  \emph{Proceedings of the Thirty-Fourth Conference on Neural Information
  Processing Systems}, 2020.

\bibitem{charoenphakdee2021focal}
N.~Charoenphakdee, J.~Vongkulbhisal, N.~Chairatanakul, and M.~Sugiyama, ``On
  focal loss for class-posterior probability estimation: {A} theoretical
  perspective,'' in \emph{Proceedings of the 2021 Conference on Computer Vision
  and Pattern Recognition}, 2021.

\bibitem{teixeira2019graph}
L.~Teixeira, B.~Jalaian, and B.~Ribeiro, ``Are graph neural networks
  miscalibrated?'' \emph{arXiv:1905.02296}, 2019.

\bibitem{wu2019simplifying}
F.~Wu, T.~Zhang, {\relax Jr}.~Souza, A. H.~d., C.~Fifty, T.~Yu, and K.~Q.
  Weinberger, ``Simplifying graph convolutional networks,'' in
  \emph{Proceedings of the Thirty-Sixth International Conference on Machine
  Learning}, 2019.

\bibitem{nt2019revisiting}
H.~NT and T.~Maehara, ``Revisiting graph neural networks: All we have is
  low-pass filters,'' \emph{arXiv:1905.09550}, 2019.

\bibitem{page1998rank}
L.~Page, S.~Brin, R.~Motwani, and T.~Winograd, ``The {PageRank} citation
  ranking: Bringing order to the web,'' \emph{Stanford InfoLab}, 1998.

\bibitem{zadrozny2001obtaining}
B.~Zadrozny and C.~Elkan, ``Obtaining calibrated probability estimates from
  decision trees and naive {B}ayesian classifiers,'' in \emph{Proceedings of
  the Eighteenth International Conference on Machine Learning}, 2001.

\bibitem{zadrozny2002transforming}
------, ``Transforming classifier scores into accurate multiclass probability
  estimates,'' in \emph{Proceedings of the Eighth ACM SIGKDD Conference on
  Knowledge Discovery and Data Mining}, 2002.

\bibitem{ma2021meta}
X.~Ma and M.~B. Blaschko, ``Meta-{Cal}: Well-controlled post-hoc calibration by
  ranking,'' in \emph{Proceedings of the Thirty-Eighth International Conference
  on Machine Learning}, 2021.

\bibitem{li2018deeper}
Q.~Li, Z.~Han, and X.-M. Wu, ``Deeper insights into graph convolutional
  networks for semi-supervised learning,'' in \emph{Proceedings of the
  Thirty-Second AAAI conference on Artificial Intelligence}, 2018.

\bibitem{xu2018representation}
K.~Xu, C.~Li, Y.~Tian, T.~Sonobe, K.-i. Kawarabayashi, and S.~Jegelka,
  ``Representation learning on graphs with jumping knowledge networks,'' in
  \emph{Proceedings of the Thirty-Fifth International Conference on Machine
  Learning}, 2018.

\end{thebibliography}

% Do not include the supplement for submission!
%\include{supplement}

\section*{Supplementary material}
\begin{table*}[b]
\setlength{\tabcolsep}{0.5em}
\setlength{\extrarowheight}{0.1em}
\caption{Calibrated accuracy (mean±SD over 100 independent runs). Temperature scaling and RBS do not change the accuracy.}
\label{table_calibration_accuracy}
\centering
\begin{tabular}{cc|cccccc}
\hline
Dataset  & Model & Uncal.        & His   & Iso     & BBQ         & Meta \\\hline
\multirow{5}{*}{Cora}     & GCN   & 81.43±0.60  & 80.38±0.82  & 81.80±0.57 &  81.34±0.67 & 79.23±1.61\\
    & GAT   & 83.14±0.39 & 81.39±0.48 &  84.05±0.51  &  83.52±0.59  & 79.99±1.70 \\
     & SGC   & 81.19±0.05  & 79.91±0.13 & 81.16±0.11 &  79.83±0.24&  78.77±1.88 \\
     & gfNN  & 78.73±5.04   & 79.06±1.16 & 80.21±1.28 &  79.96±1.31 & 76.30±5.51\\
   & APPNP &  83.68±0.36 & 82.52±0.46 & 83.45±0.45 &  83.20±0.53&81.33±1.79 \\\hline
\multirow{5}{*}{Citeseer}  & GCN   & 71.32±0.70    &71.93±0.84 &   72.39±0.66& 71.79±0.99 & 68.22±4.13\\
 & GAT   &  70.99±0.60  & 71.78±0.56 &  72.21±0.52 & 71.81±0.70 & 68.28±2.63 \\
 & SGC   & 72.46±0.15  & 74.13±0.05 & 73.81±0.12&  73.32±0.13  &  69.19±2.08\\
 & gfNN  & 67.33±6.58   &71.74±1.22 &71.98±1.15  & 71.10±1.22 & 64.63±6.61\\
& APPNP & 72.10±0.38  & 72.94±0.48 & 72.90±0.48& 72.63±0.83 & 69.64±2.29 \\\hline
\multirow{5}{*}{Pubmed}   & GCN   & 79.23±0.43 & 79.01±0.55  & 79.03±0.46& 78.85±0.67 & 76.99±4.62 \\
  & GAT   & 79.05±0.38   & 78.50±0.61 & 78.85±0.38& 78.19±0.56 & 78.00±1.43 \\
 &SGC & 78.72±0.04  & 79.05±0.08  & 79.30±0.03&   79.88±0.19 & 77.83±1.57\\
   & gfNN  & 77.94±2.32  &  77.92±1.11 &78.16±0.98 &  77.76±1.33  &  75.66±3.05\\
   & APPNP & 80.09±0.25  & 80.12±0.44 &  80.15±0.30& 79.50±0.47   & 78.37±1.64 \\\hline
\end{tabular}
\end{table*}

\textit{GNN models} 
We follow the best hyperparameter and training settings given by the corresponding graph neural network papers, using the PyTorch-Geometric implementation\footnote{\url{https://github.com/pyg-team/pytorch_geometric/tree/master/benchmark/citation}} of GCN~\cite{kipf2016semi}, GAT~\cite{velivckovic2017graph}, SGC~\cite{wu2019simplifying}, and APPNP~\cite{klicpera2019predict}.
For SGC, we set the learning rate to 0.2, train for 100 epochs without early stopping, and tune weight decay for 60 iterations using Hyperopt\footnote{\url{https://github.com/hyperopt/hyperopt}}, according to the original paper.
We implement gfNN~\cite{nt2019revisiting} by ourselves and follow the setting in the original paper. 
In order to capture the variance across different training runs, each model is run for 100 times, and we report the averaged results with standard deviations.\\

\noindent \textit{Width} We conduct experiments on the influence of width on the models GCN and GAT, where we use the best hyperparameter settings and vary the hidden dimensions per layer in the range given by  $\{2^{i} \mid 3\leq i \leq 10\}$. Dropout layers are removed, and the number of epochs is set to 200 with early stopping after 10 epochs without improvement of the validation loss. Each model is run for 10 times.\\

\noindent \textit{Depth} We conduct experiments on the influence of depth on the models GCN and GAT, where we follow the experimental setting in Appendix B by Kipf and Welling~\cite{kipf2016semi}. The number of layers is in the range $\{1, 2, \dots, 10\}$. Each model is run for 10 times.\\
%The hidden dimension is 16, and we train for 400 epochs (without early stopping) using Adam (with learning rate 0.01 and weight decay 5e-4); dropout rate = 0.5, in the first and last layer.

\noindent \textit{Graph density}
We conduct experiments on the influence of graph density on the models GCN and GAT. %, where we use the same hyperparameter settings for both models. The number of attention heads for GAT is set to 8, which is consistent with the best hyperparameter setting. 
Different proportions of edges are removed randomly from $0\%$ (original dataset) to $100\%$ (no graph structure at all). Each model is run for 10 times.\\

\noindent \textit{New loss function}
We follow the setting by Tomani and Buettner~\cite{tomani2021trustworthy}, who also introduced an ECE-inspired loss function. An annealing coefficient is specified for the calibration error term since the early epochs are usually used for reaching the cross entropy minimum. More precisely, we define
\begin{align}
\mathrm{anneal\_coef} &=   \lambda \cdot \min\left\{1, \frac{\mathrm{epoch}}{\mathrm{EPOCHS} \cdot \mathrm{anneal\_max}}\right\}\,, \\
\tilde{L}_{\mathrm{cal}} &= \mathrm{anneal\_coef} \cdot L_{\mathrm{cal}}\,,\quad \mathrm{and}\\
L &= \alpha \cdot L_{\mathrm{ce}} + (1- \alpha) \cdot \tilde{L}_{\mathrm{cal}}\;,
\end{align}
where epoch and EPOCHS are the current training epoch and the total number of epochs, respectively, and $\lambda$, $\mathrm{anneal\_max}$, and $\alpha$ are hyperparameters.
We set $\mathrm{anneal\_max} = 1$, $\lambda = 1$, and tune $\alpha$ on the validation set in the range $\{ 0.95, 0.96, 0.97, 0.98, 0.99\}$.
%The best hyperparameters are displayed in Table~\ref{tab:loss}. 
Each model (fixed hyperparameter setting) is run for 10 times.\\

\begin{figure*}[t]
\centering 
\subfigure{
\includegraphics[width=0.25\textwidth]{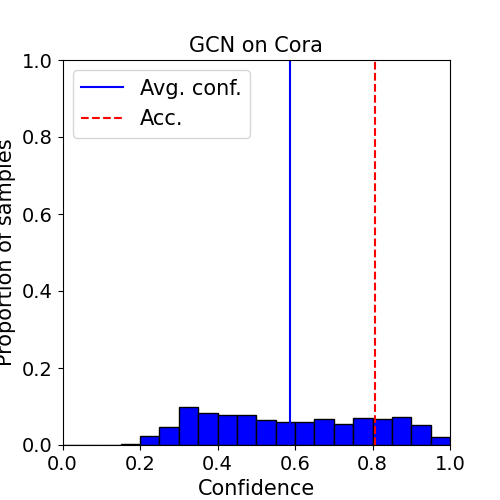}}\hspace{-4mm}
\subfigure{
\includegraphics[width=0.25\textwidth]{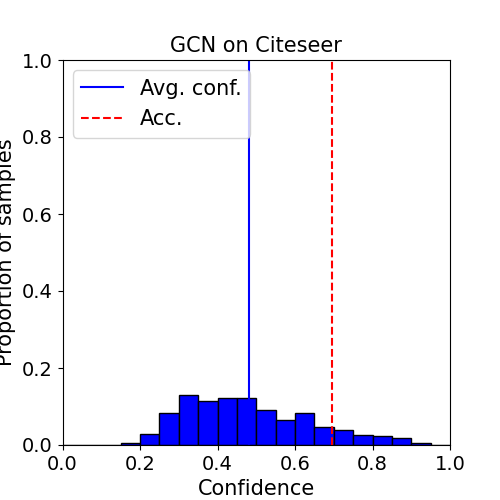}}\hspace{-4mm}
\subfigure{
\includegraphics[width=0.25\textwidth]{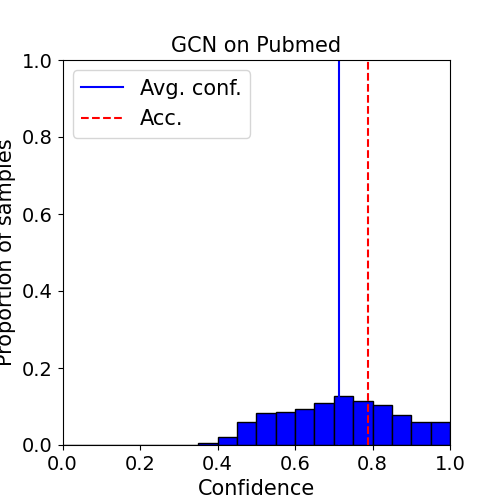}}\\
\subfigure{
\includegraphics[width=0.25\textwidth]{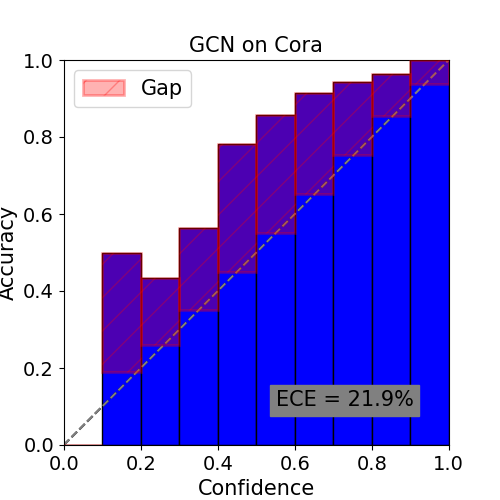}}\hspace{-4mm}
\subfigure{
\includegraphics[width=0.25\textwidth]{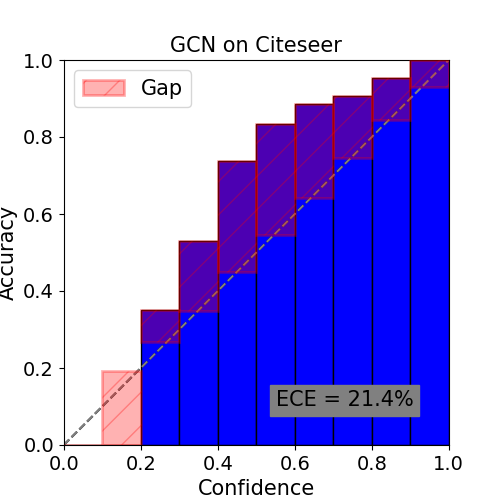}}\hspace{-4mm}
\subfigure{
\includegraphics[width=0.25\textwidth]{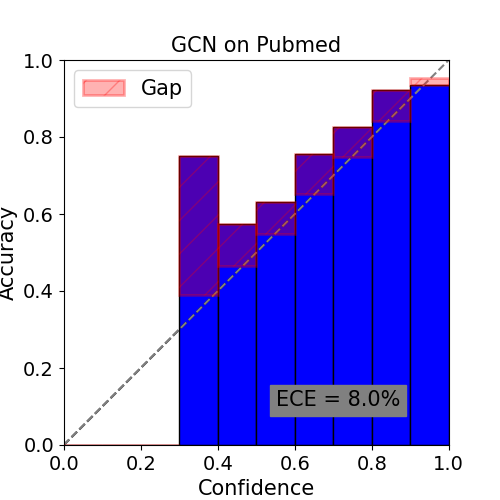}}\\
\subfigure{
\includegraphics[width=0.25\textwidth]{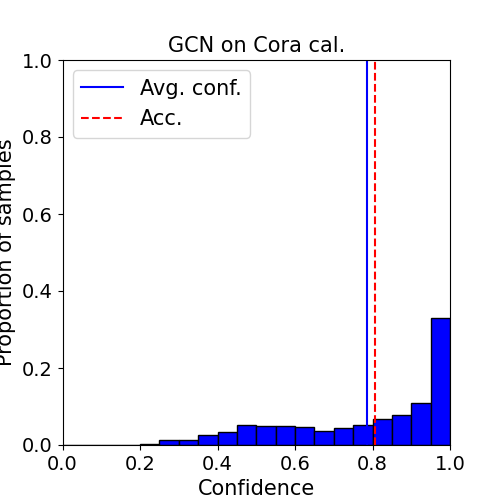}}\hspace{-4mm}
\subfigure{
\includegraphics[width=0.25\textwidth]{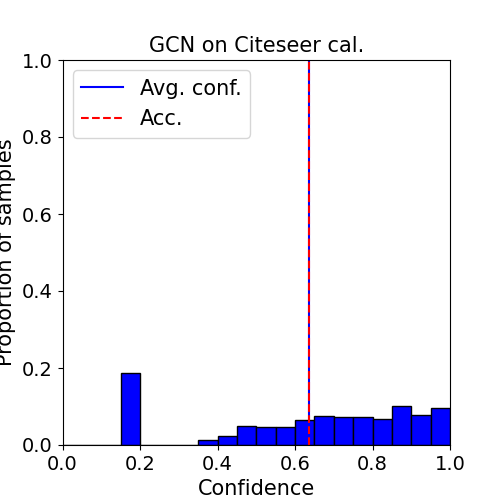}}\hspace{-4mm}
\subfigure{
\includegraphics[width=0.25\textwidth]{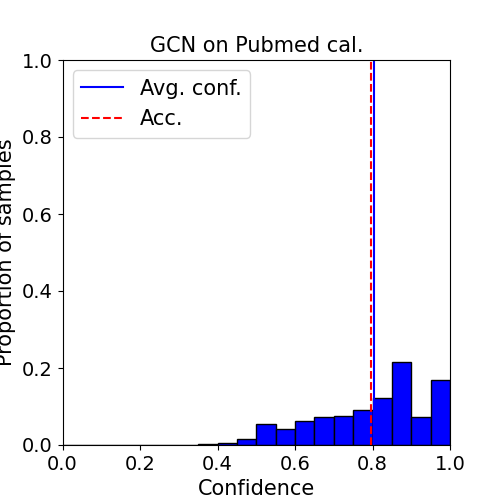}}\\
\subfigure{
\includegraphics[width=0.25\textwidth]{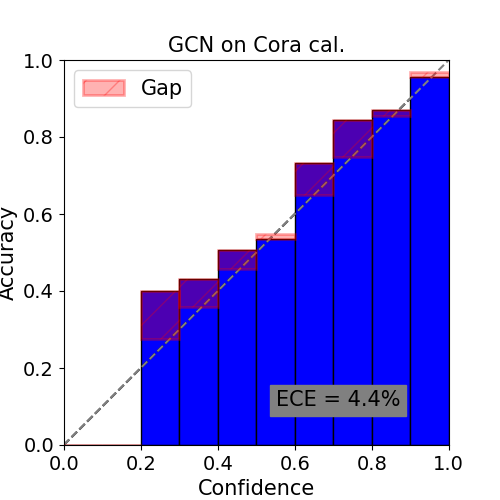}}\hspace{-4mm}
\subfigure{
\includegraphics[width=0.25\textwidth]{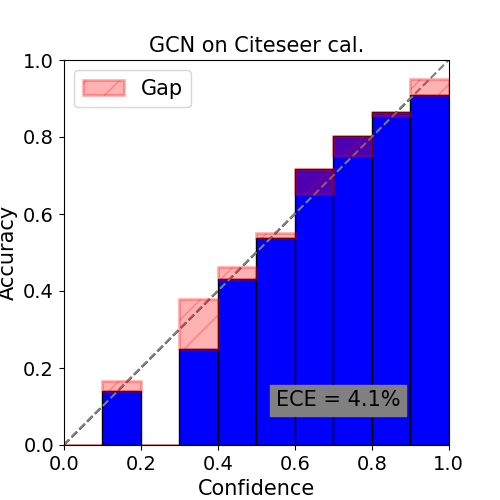}}\hspace{-4mm}
\subfigure{
\includegraphics[width=0.25\textwidth]{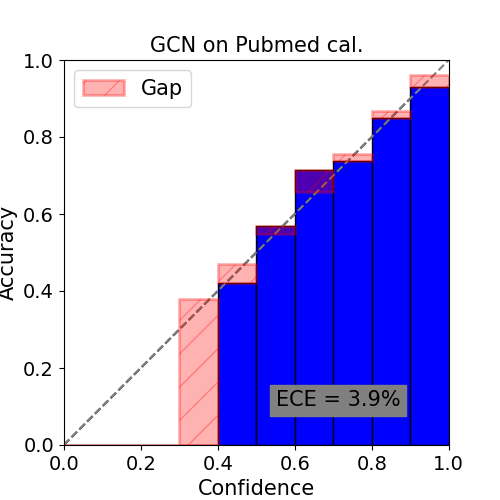}}\\
\caption{Histograms and reliability diagrams for GCN.}
\end{figure*}

\begin{figure*}[t]
\centering 
\subfigure{
\includegraphics[width=0.25\textwidth]{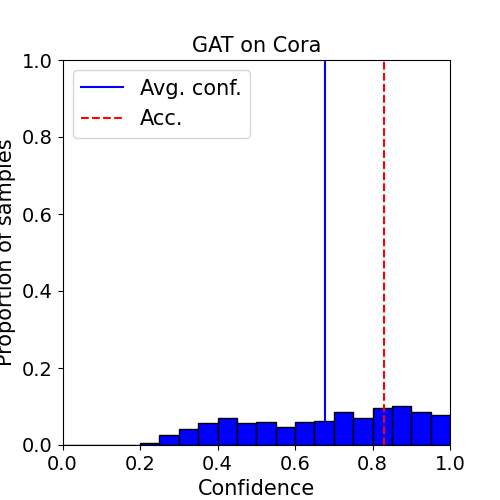}}\hspace{-4mm}
\subfigure{
\includegraphics[width=0.25\textwidth]{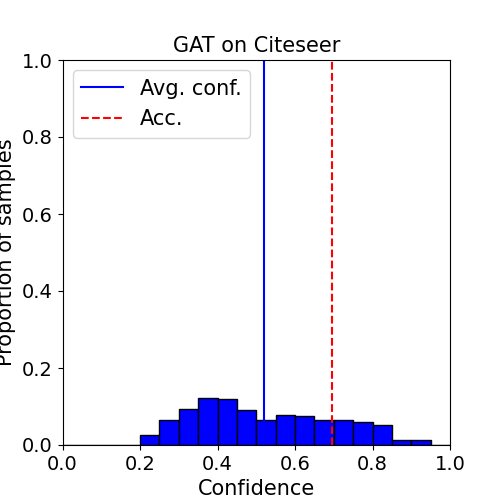}}\hspace{-4mm}
\subfigure{
\includegraphics[width=0.25\textwidth]{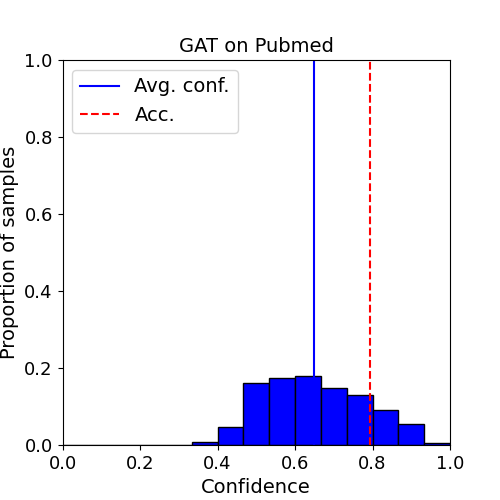}}\\
\subfigure{
\includegraphics[width=0.25\textwidth]{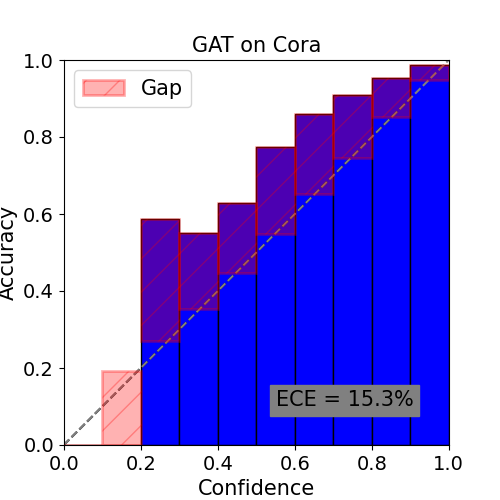}}\hspace{-4mm}
\subfigure{
\includegraphics[width=0.25\textwidth]{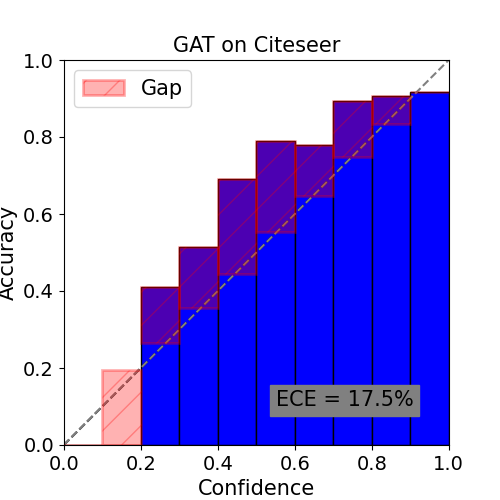}}\hspace{-4mm}
\subfigure{
\includegraphics[width=0.25\textwidth]{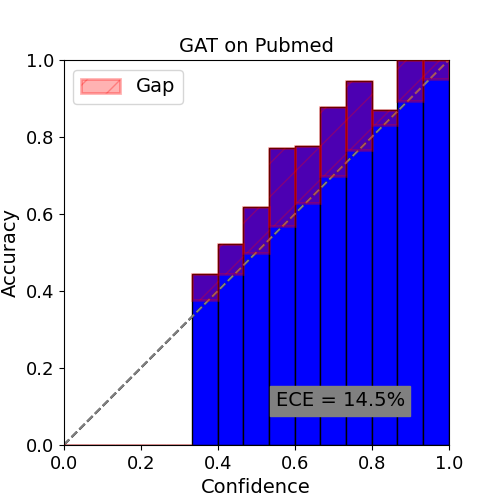}}\\
\subfigure{
\includegraphics[width=0.25\textwidth]{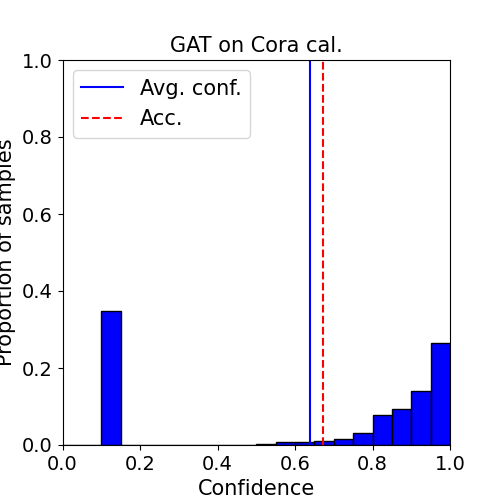}}\hspace{-4mm}
\subfigure{
\includegraphics[width=0.25\textwidth]{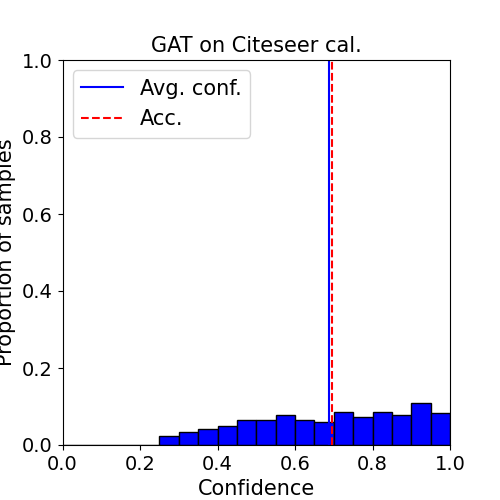}}\hspace{-4mm}
\subfigure{
\includegraphics[width=0.25\textwidth]{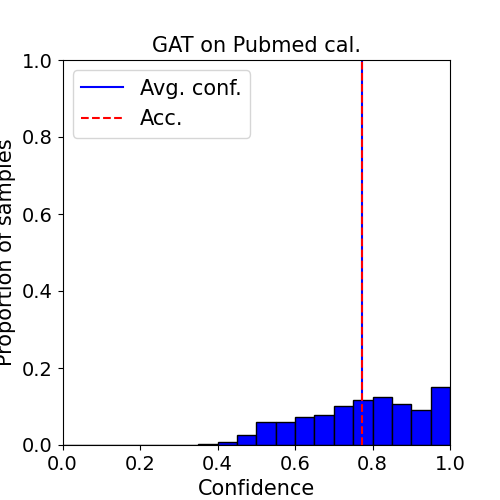}}\\
\subfigure{
\includegraphics[width=0.25\textwidth]{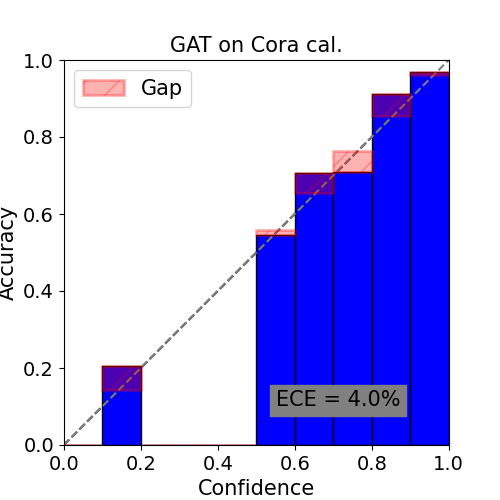}}\hspace{-4mm}
\subfigure{
\includegraphics[width=0.25\textwidth]{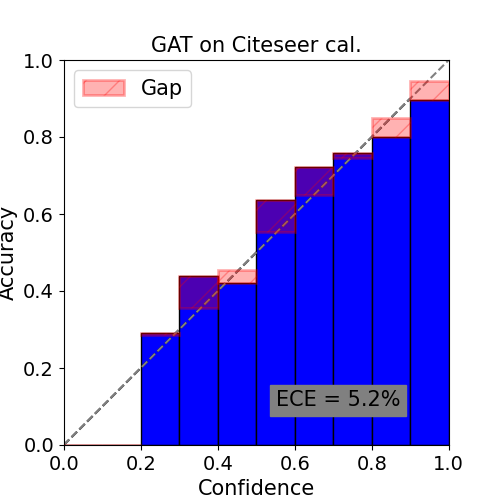}}\hspace{-4mm}
\subfigure{
\includegraphics[width=0.25\textwidth]{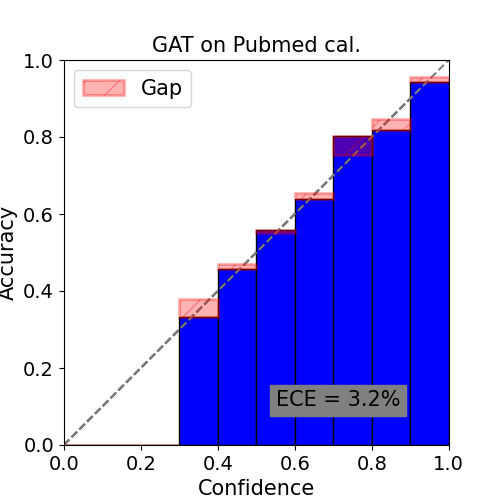}}
\caption{Histograms and reliability diagrams for GAT.}
\end{figure*}

\begin{figure*}[t]
\centering 
\subfigure{
\includegraphics[width=0.25\textwidth]{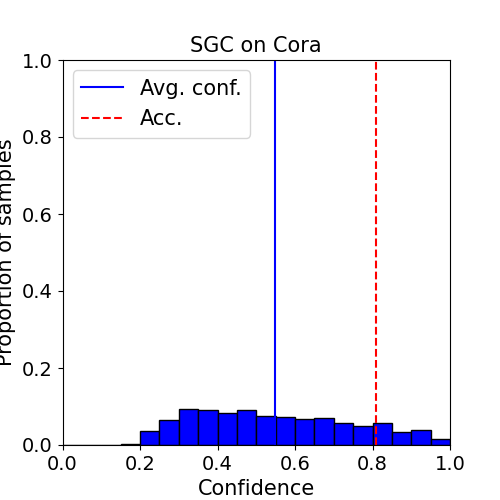}}\hspace{-4mm}
\subfigure{
\includegraphics[width=0.25\textwidth]{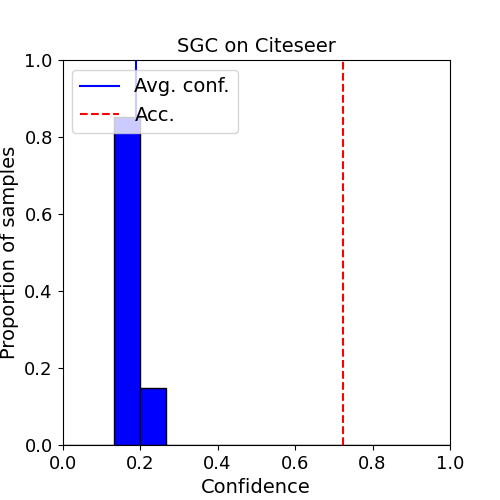}}\hspace{-4mm}
\subfigure{
\includegraphics[width=0.25\textwidth]{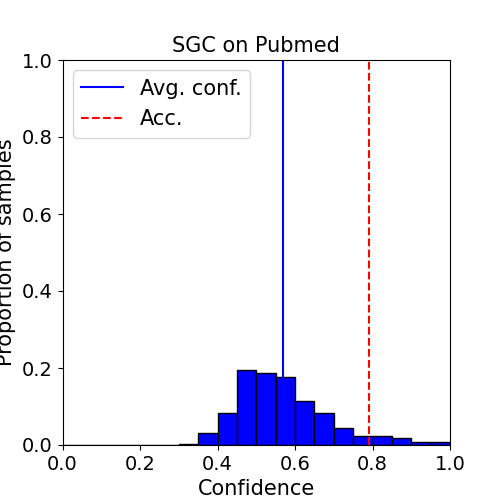}}\\
\subfigure{
\includegraphics[width=0.25\textwidth]{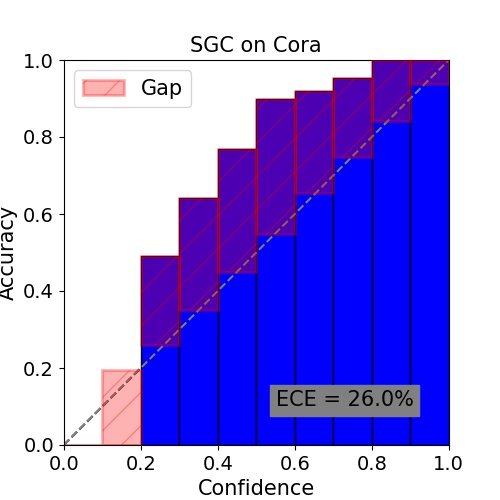}}\hspace{-4mm}
\subfigure{
\includegraphics[width=0.25\textwidth]{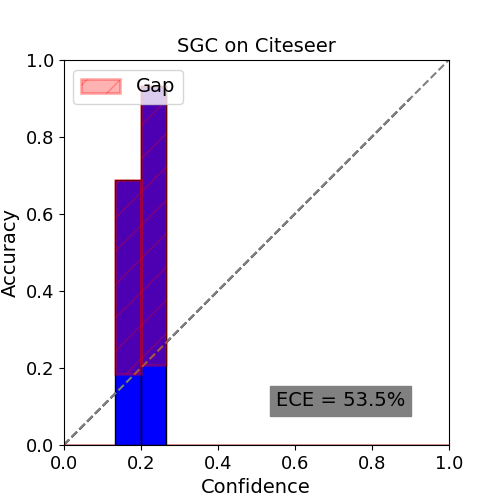}}\hspace{-4mm}
\subfigure{
\includegraphics[width=0.25\textwidth]{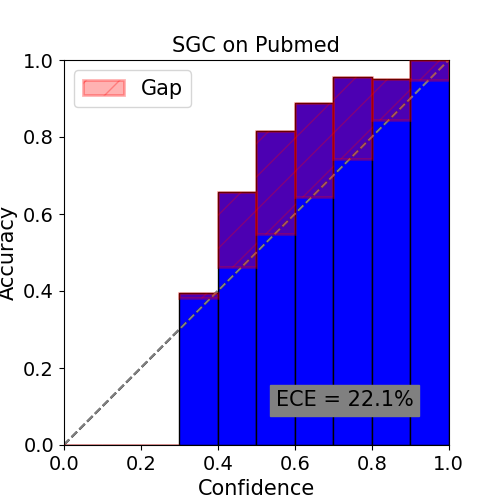}}\\
\subfigure{
\includegraphics[width=0.25\textwidth]{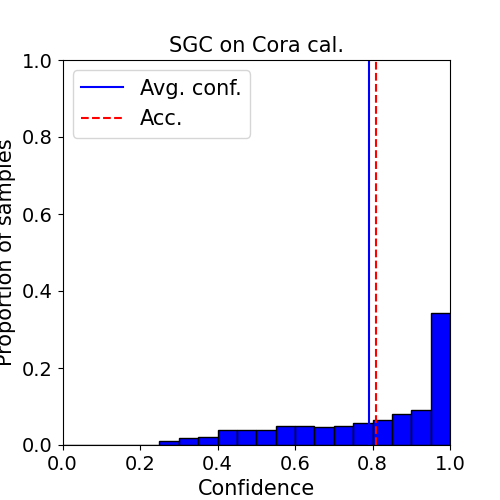}}\hspace{-4mm}
\subfigure{
\includegraphics[width=0.25\textwidth]{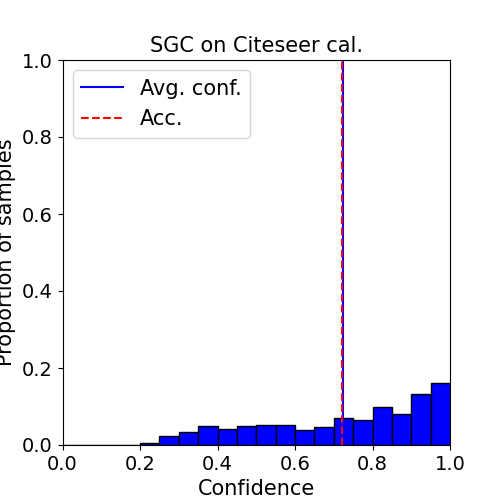}}\hspace{-4mm}
\subfigure{
\includegraphics[width=0.25\textwidth]{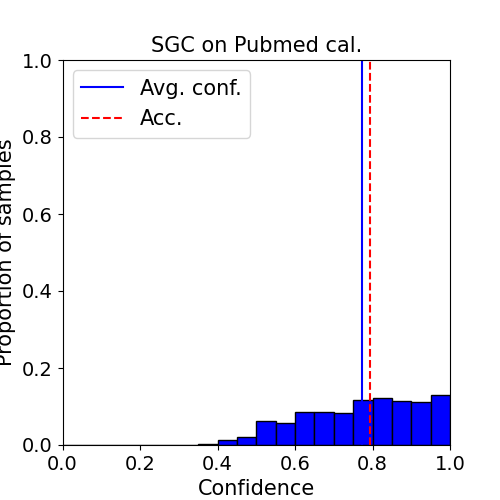}}\\
\subfigure{
\includegraphics[width=0.25\textwidth]{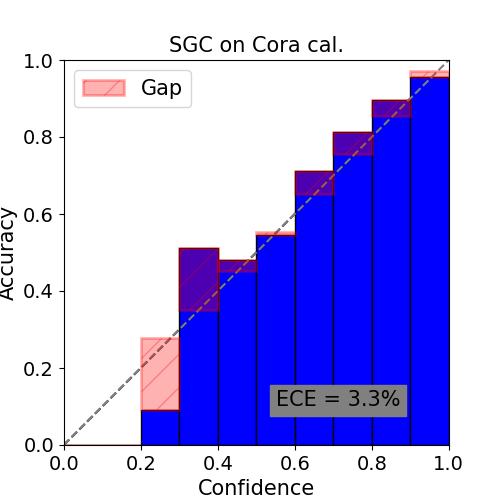}}\hspace{-4mm}
\subfigure{
\includegraphics[width=0.25\textwidth]{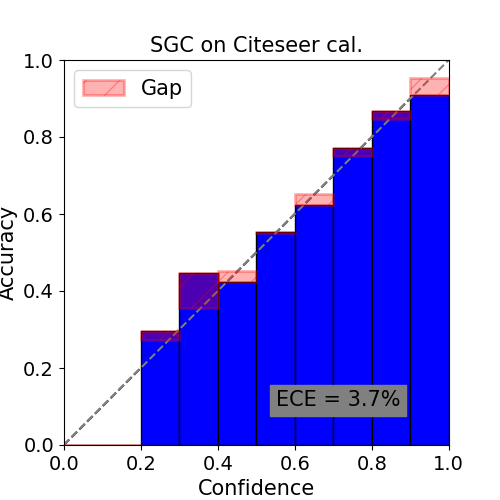}}\hspace{-4mm}
\subfigure{
\includegraphics[width=0.25\textwidth]{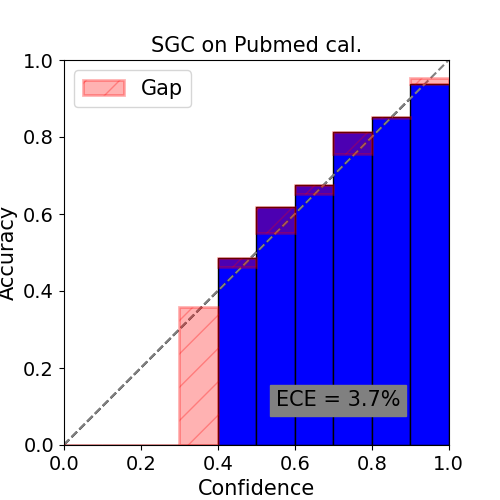}}
\caption{Histograms and reliability diagrams for SGC.}
\end{figure*}

\begin{figure*}[t]
\centering 
\subfigure{
\includegraphics[width=0.25\textwidth]{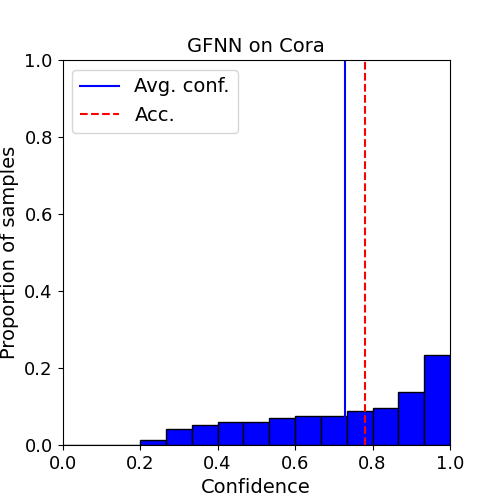}}\hspace{-4mm}
\subfigure{
\includegraphics[width=0.25\textwidth]{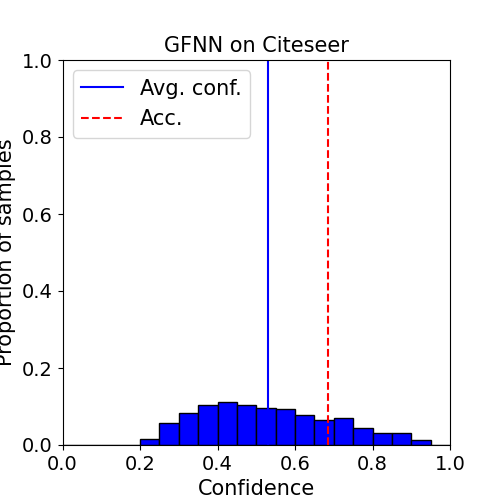}}\hspace{-4mm}
\subfigure{
\includegraphics[width=0.25\textwidth]{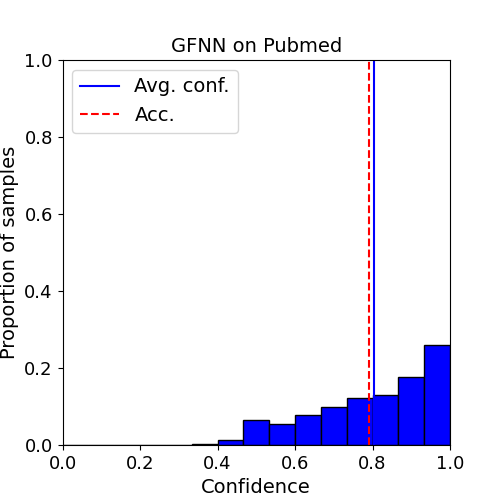}}\\
\subfigure{
\includegraphics[width=0.25\textwidth]{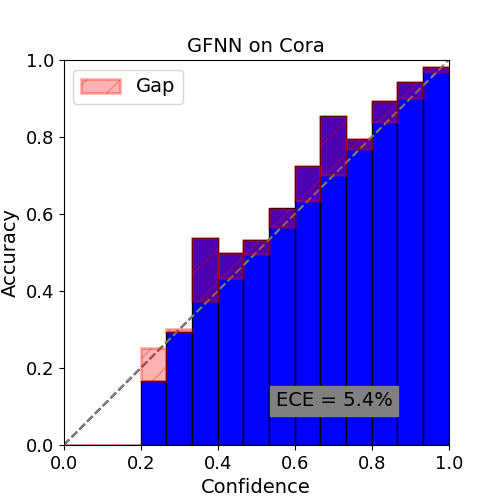}}\hspace{-4mm}
\subfigure{
\includegraphics[width=0.25\textwidth]{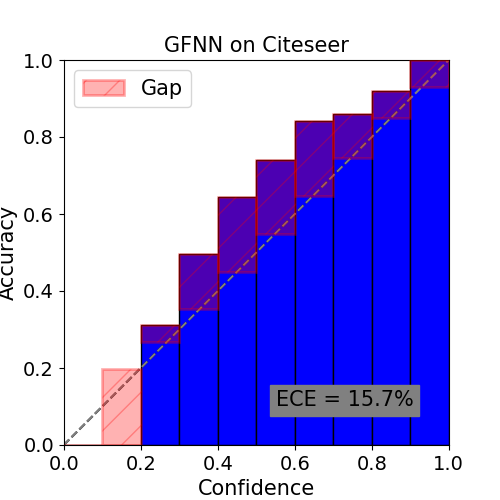}}\hspace{-4mm}
\subfigure{
\includegraphics[width=0.25\textwidth]{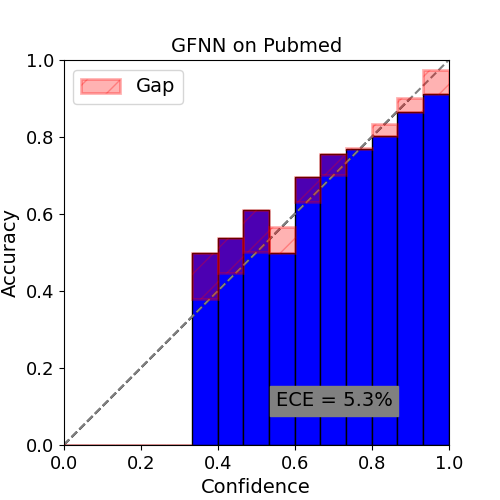}}\\
\subfigure{
\includegraphics[width=0.25\textwidth]{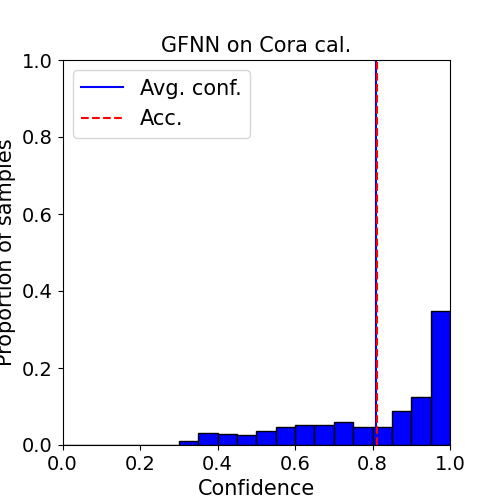}}\hspace{-4mm}
\subfigure{
\includegraphics[width=0.25\textwidth]{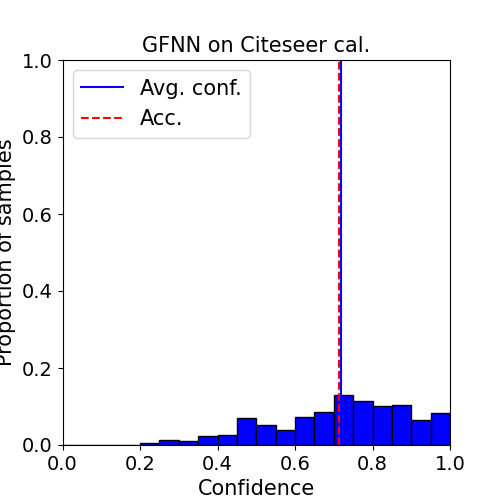}}\hspace{-4mm}
\subfigure{
\includegraphics[width=0.25\textwidth]{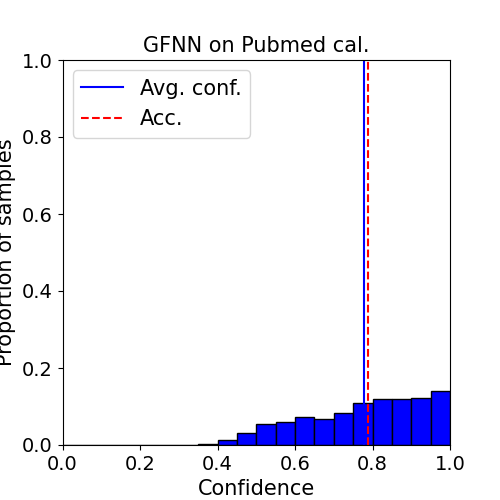}}\\
\subfigure{
\includegraphics[width=0.25\textwidth]{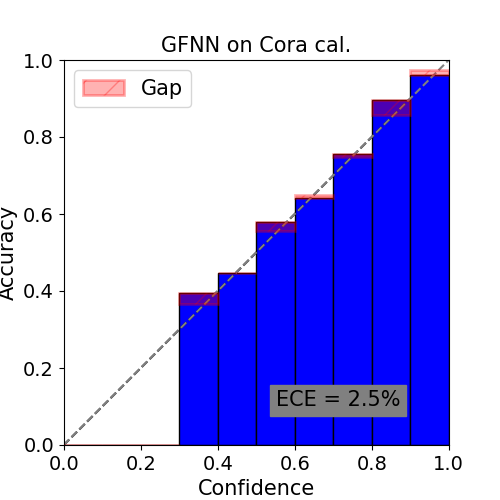}}\hspace{-4mm}
\subfigure{
\includegraphics[width=0.25\textwidth]{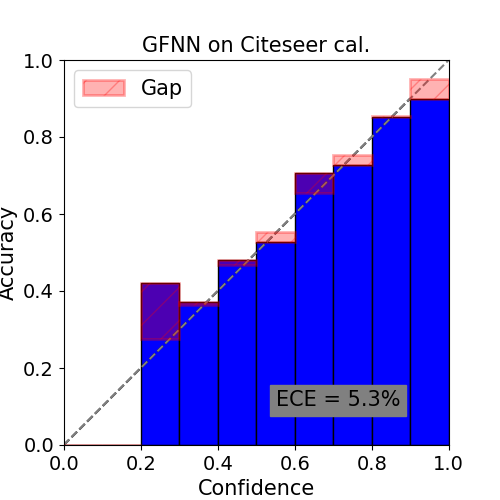}}\hspace{-4mm}
\subfigure{
\includegraphics[width=0.25\textwidth]{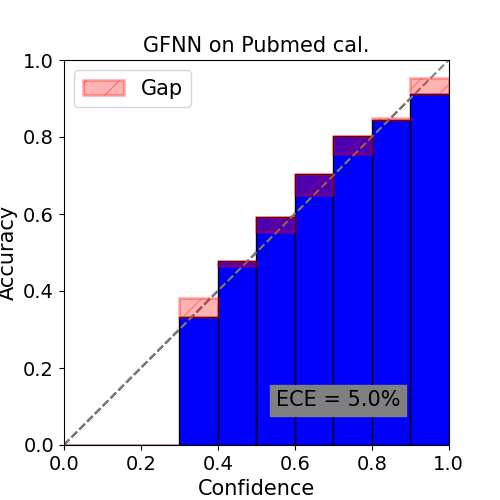}}\\
\caption{Histograms and reliability diagrams for gfNN.}
\end{figure*}

\begin{figure*}[t]
\centering 
\subfigure{
\includegraphics[width=0.25\textwidth]{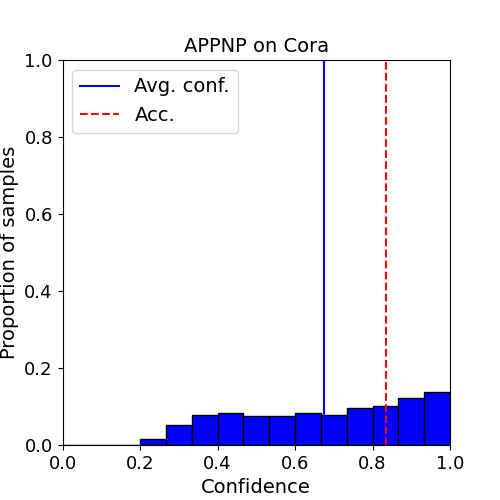}}\hspace{-4mm}
\subfigure{
\includegraphics[width=0.25\textwidth]{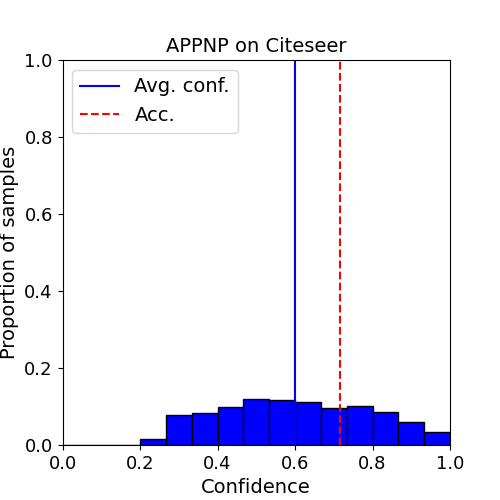}}\hspace{-4mm}
\subfigure{
\includegraphics[width=0.25\textwidth]{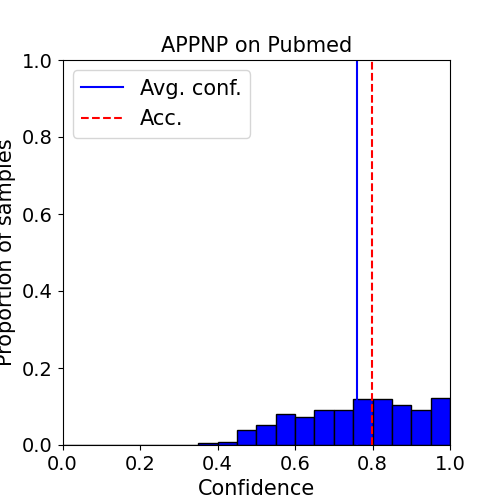}}\\
\subfigure{
\includegraphics[width=0.25\textwidth]{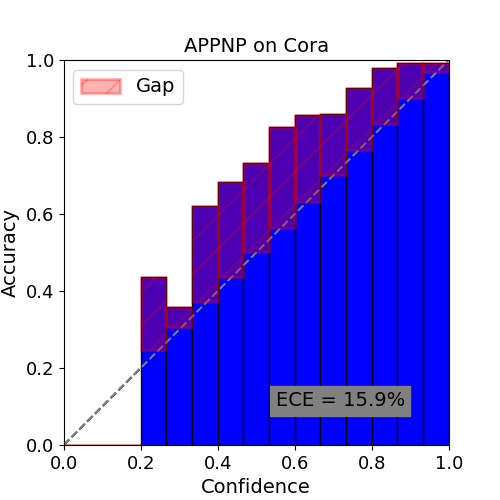}}\hspace{-4mm}
\subfigure{
\includegraphics[width=0.25\textwidth]{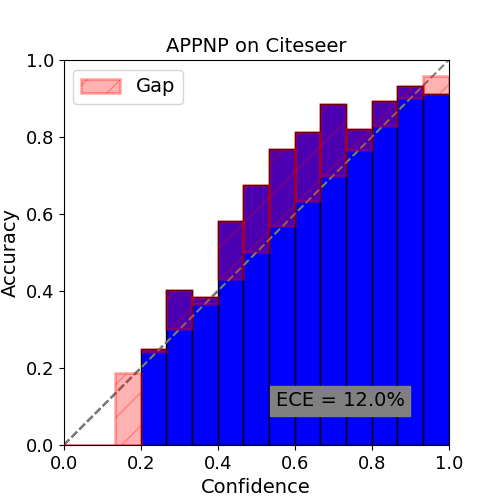}}\hspace{-4mm}
\subfigure{
\includegraphics[width=0.25\textwidth]{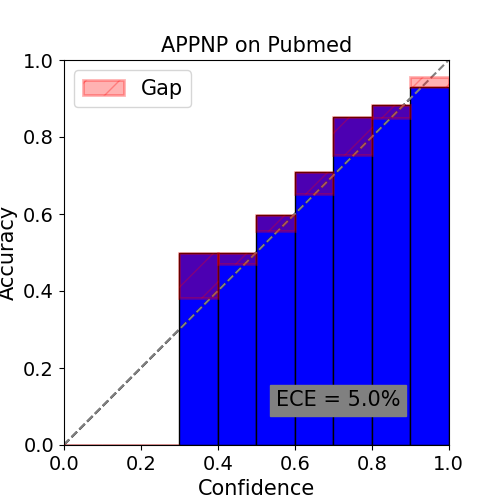}}\\
\subfigure{
\includegraphics[width=0.25\textwidth]{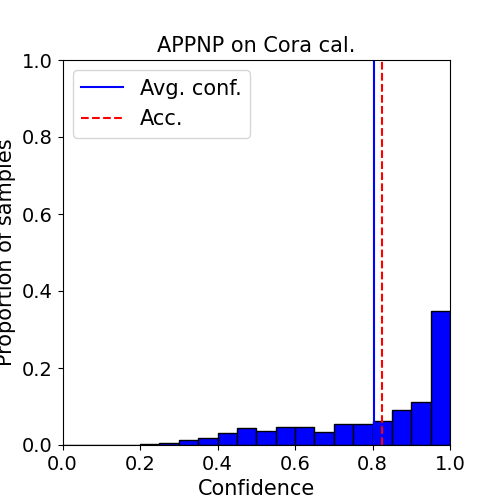}}\hspace{-4mm}
\subfigure{
\includegraphics[width=0.25\textwidth]{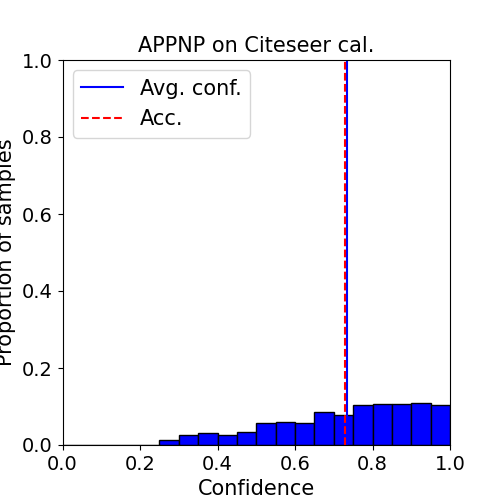}}\hspace{-4mm}
\subfigure{
\includegraphics[width=0.25\textwidth]{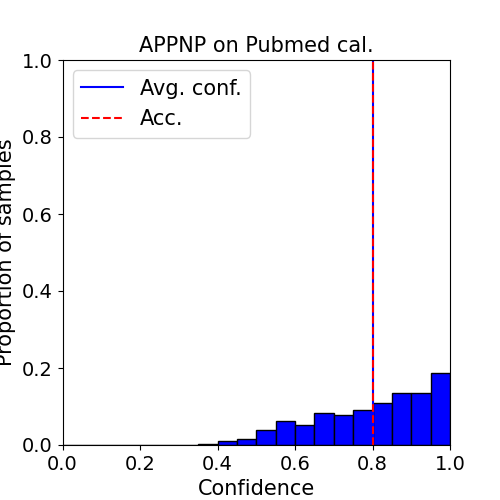}}\\
\subfigure{
\includegraphics[width=0.25\textwidth]{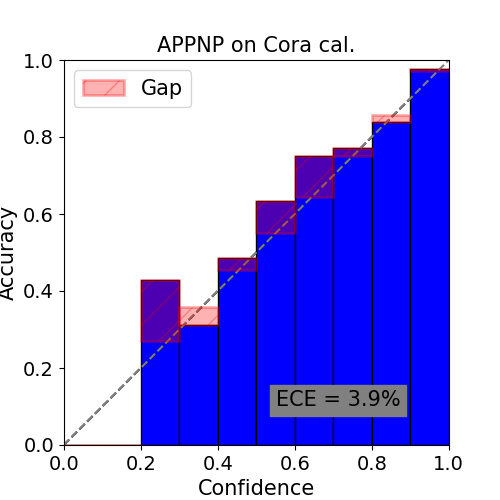}}\hspace{-4mm}
\subfigure{
\includegraphics[width=0.25\textwidth]{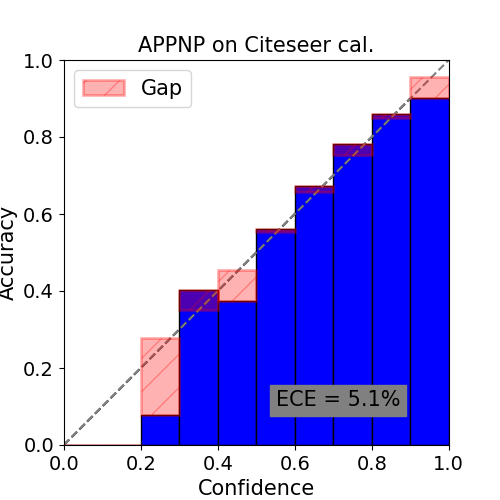}}\hspace{-4mm}
\subfigure{
\includegraphics[width=0.25\textwidth]{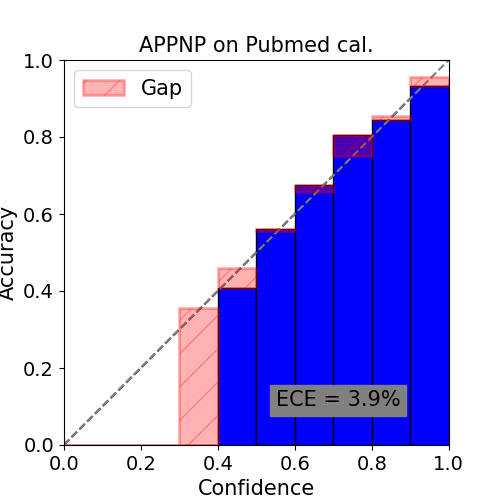}}\\
\caption{Histograms and reliability diagrams for APPNP.}
\end{figure*}

\begin{figure*}[t]
\centering 
\subfigure{
\includegraphics[width=0.25\textwidth]{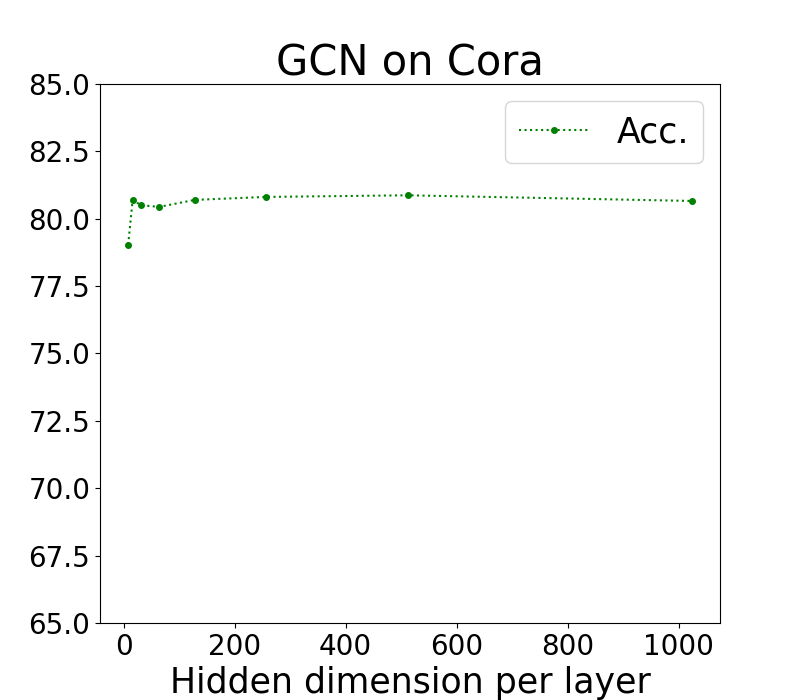}}\hspace{-6mm}
\subfigure{
\includegraphics[width=0.25\textwidth]{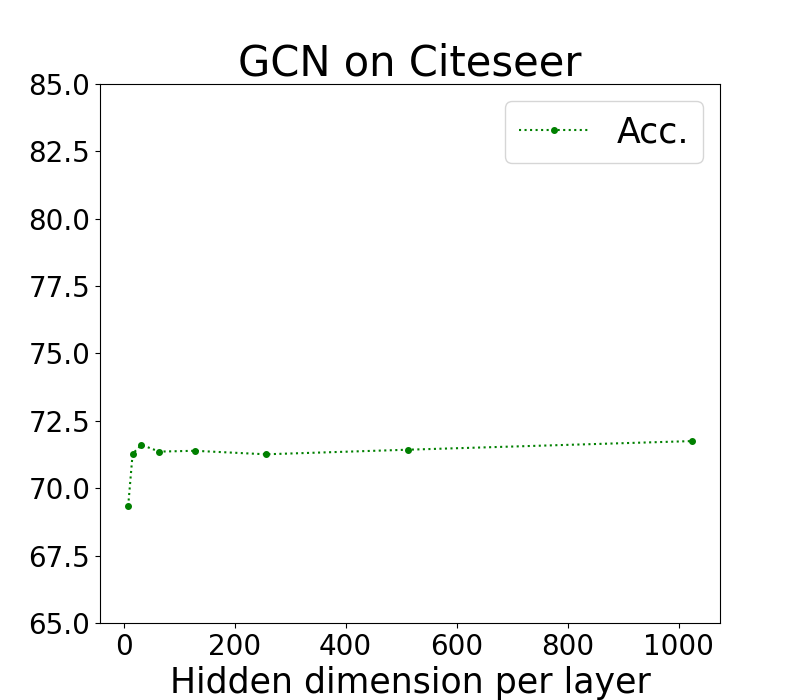}}\hspace{-6mm}
\subfigure{
\includegraphics[width=0.25\textwidth]{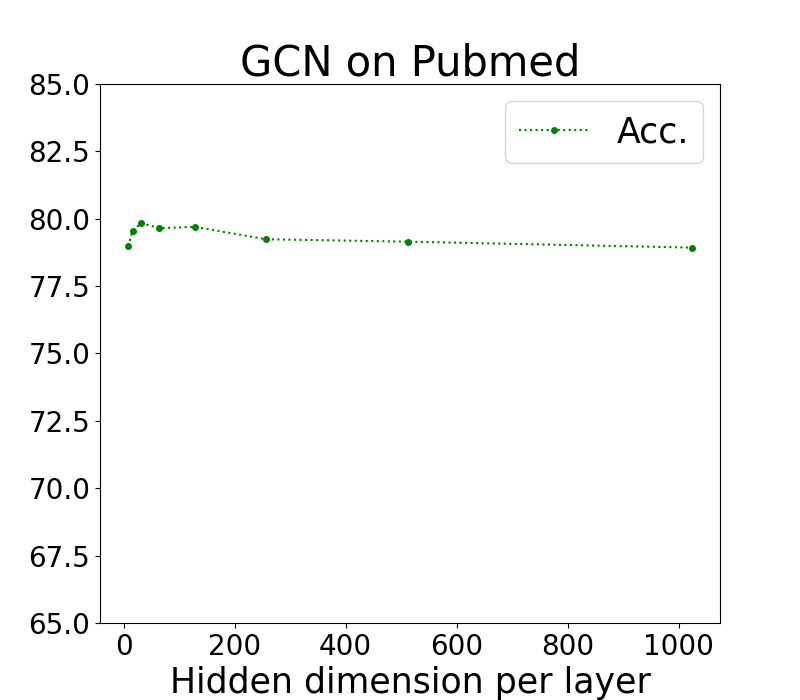}}\\
\subfigure{
\includegraphics[width=0.25\textwidth]{figures/Width/width_gcn_Cora0.png}}\hspace{-6mm}
\subfigure{
\includegraphics[width=0.25\textwidth]{figures/Width/width_gcn_Citeseer0.png}}\hspace{-6mm}
\subfigure{
\includegraphics[width=0.25\textwidth]{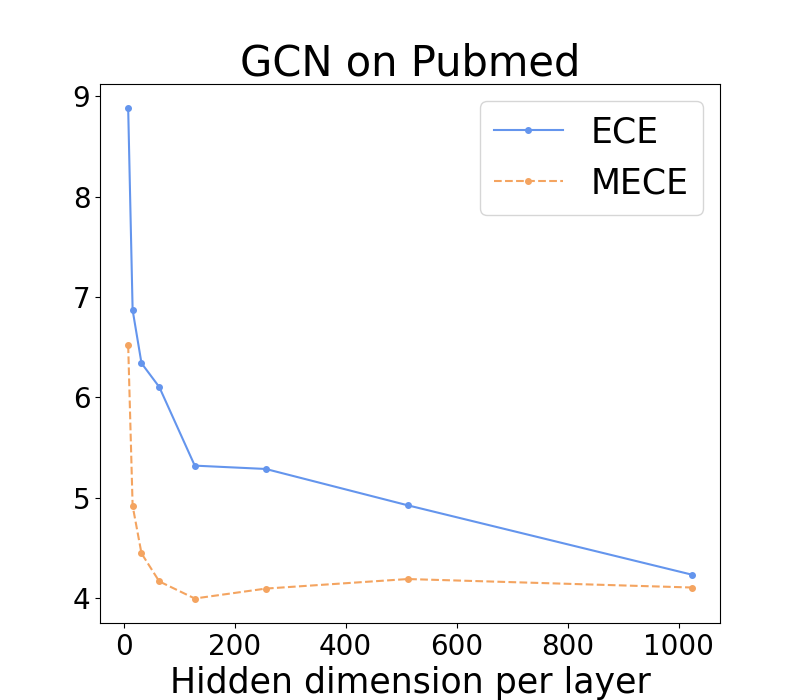}}\\
\subfigure{
\includegraphics[width=0.25\textwidth]{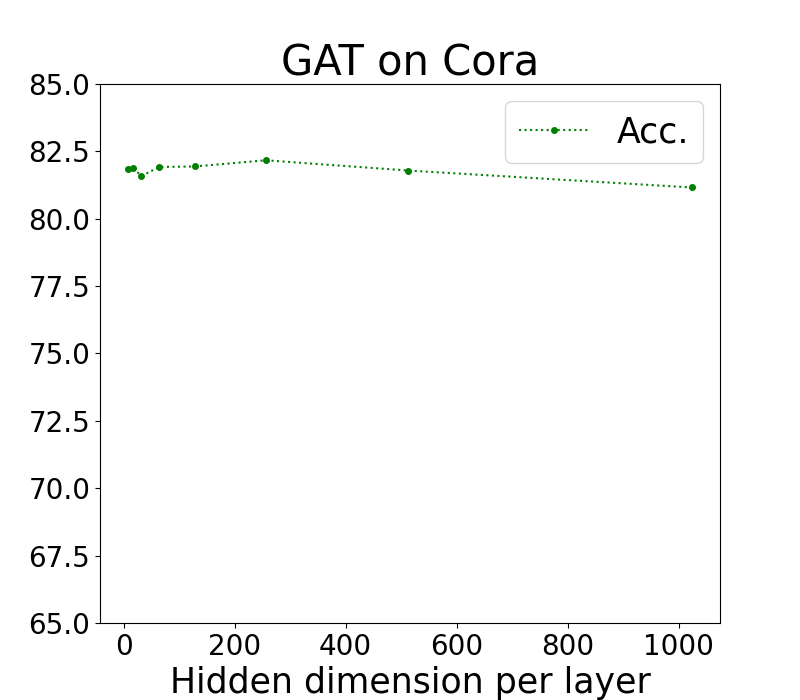}}\hspace{-6mm}
\subfigure{
\includegraphics[width=0.25\textwidth]{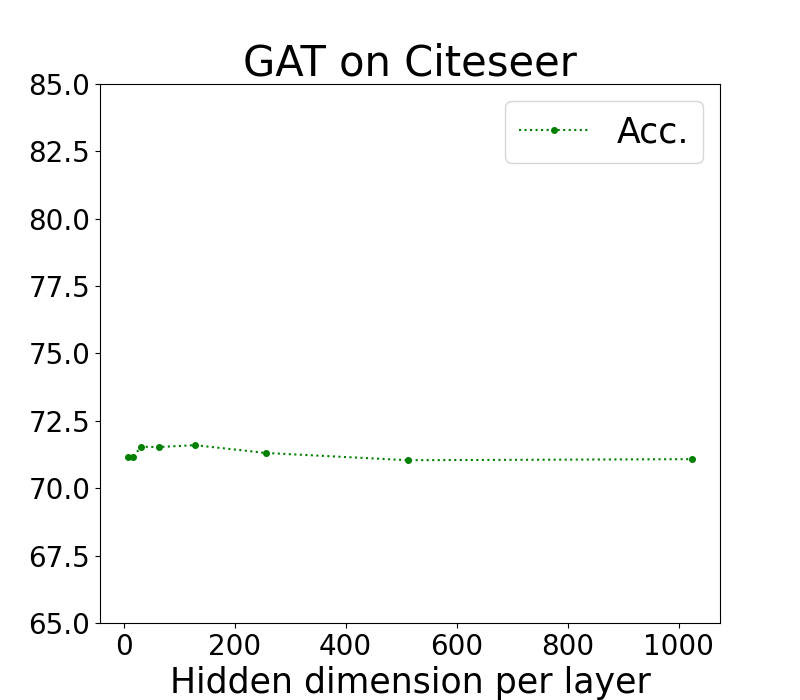}}\hspace{-6mm}
\subfigure{
\includegraphics[width=0.254\textwidth]{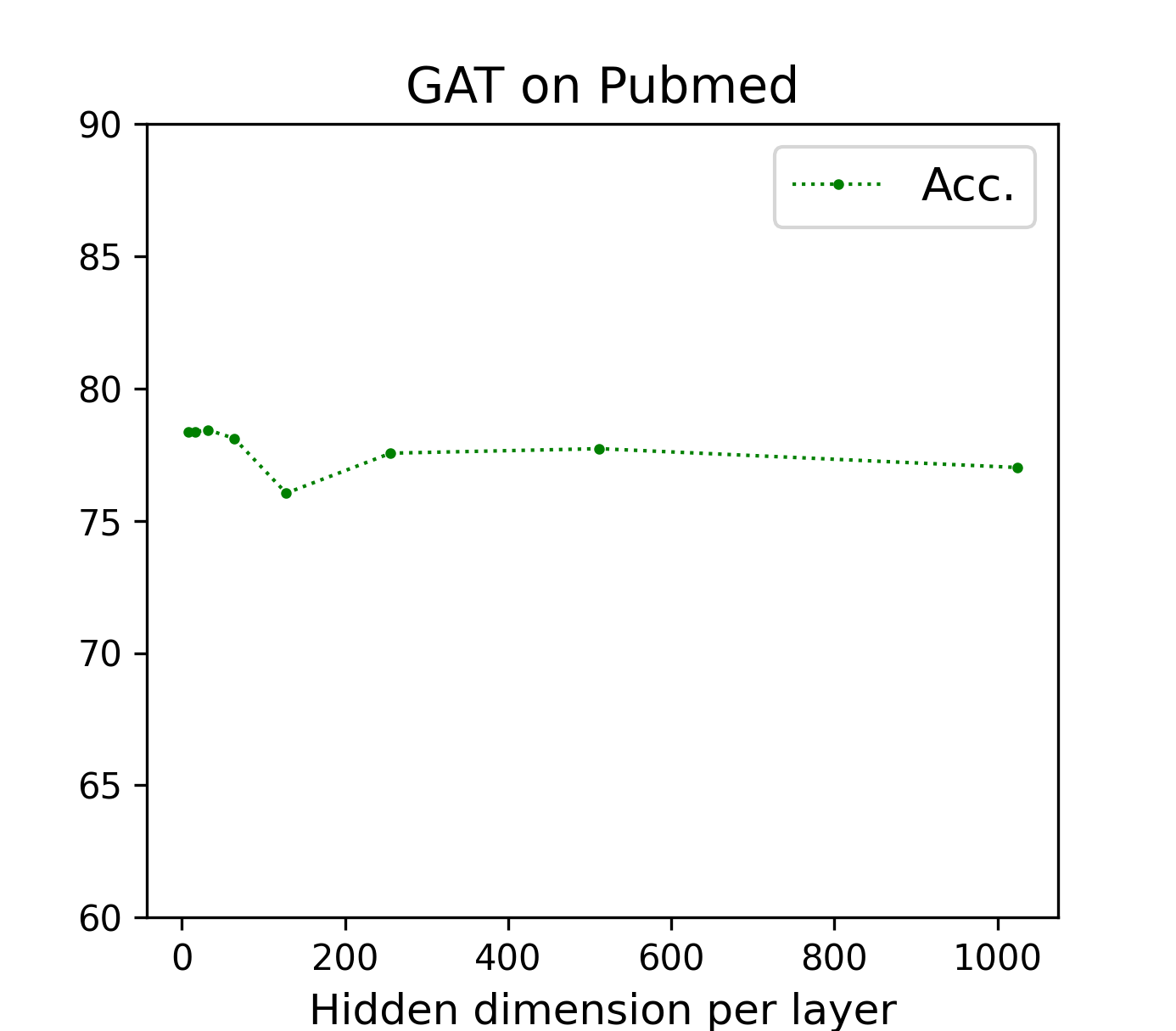}\vspace{1cm}}
\\
\subfigure{
\includegraphics[width=0.25\textwidth]{figures/Width/width_gat_Cora0.png}}\hspace{-6mm}
\subfigure{
\includegraphics[width=0.25\textwidth]{figures/Width/width_gat_Citeseer0.png}}\hspace{-6mm}
\subfigure{
\includegraphics[width=0.254\textwidth]{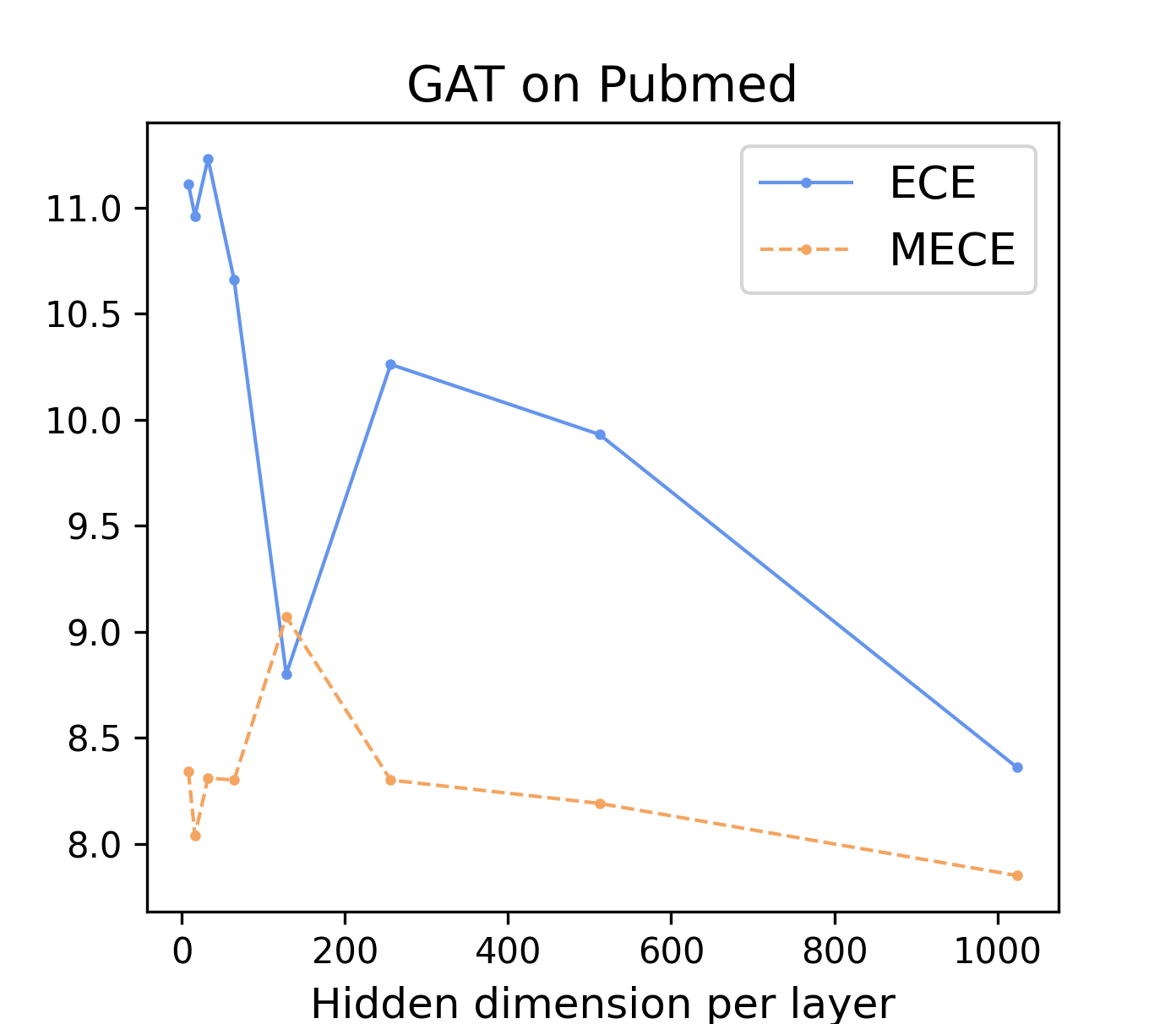}}
\caption{Influence of width.}
\end{figure*}

\begin{figure*}[t]
\centering 
\subfigure{
\includegraphics[width=0.25\textwidth]{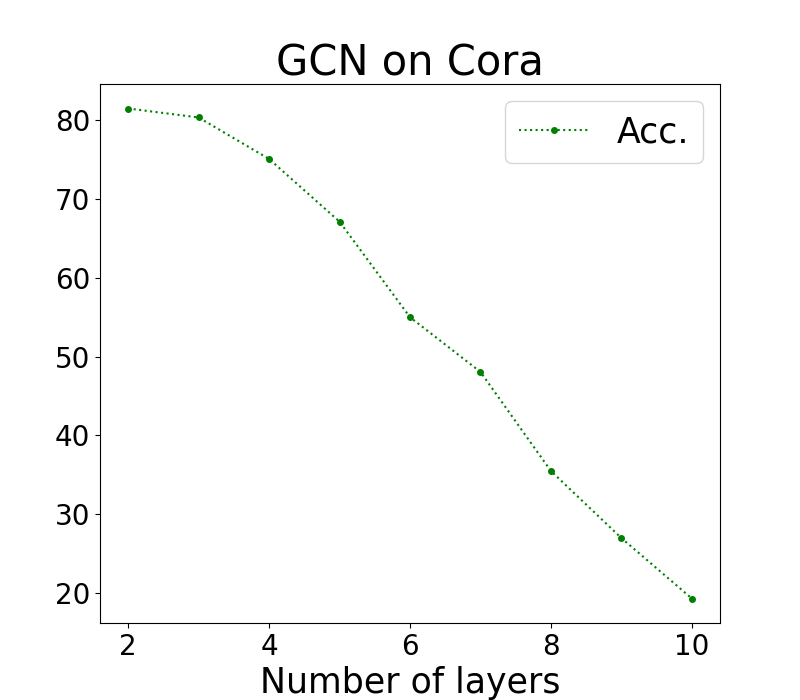}}\hspace{-6mm}
\subfigure{
\includegraphics[width=0.25\textwidth]{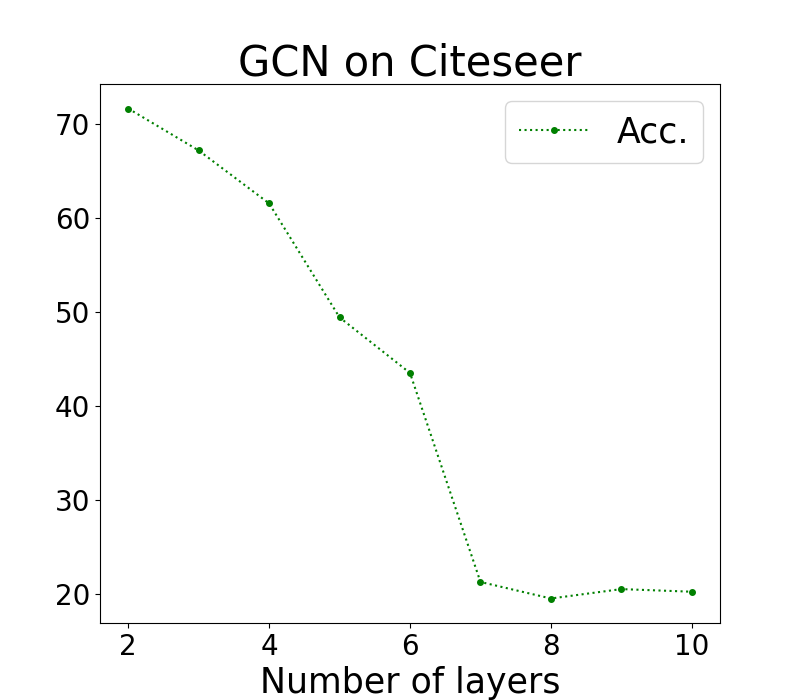}}\hspace{-6mm}
\subfigure{
\includegraphics[width=0.25\textwidth]{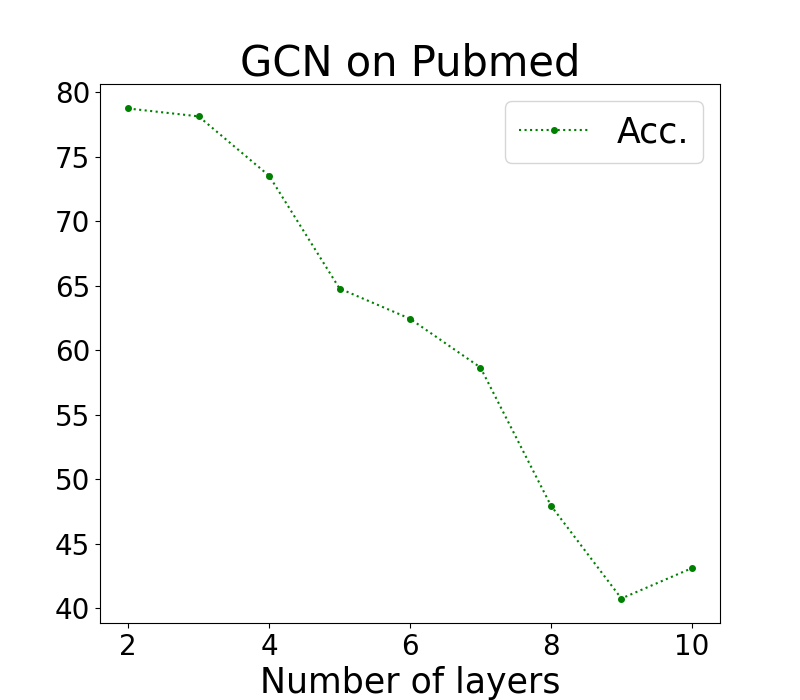}}\\
\subfigure{
\includegraphics[width=0.25\textwidth]{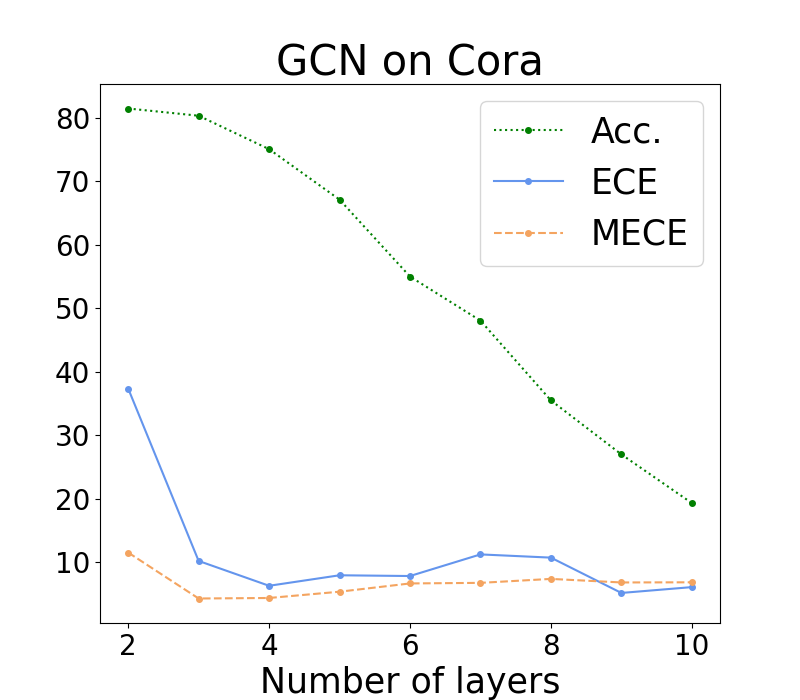}}\hspace{-6mm}
\subfigure{
\includegraphics[width=0.25\textwidth]{figures/Depth/depth_gcn_Citeseer0.png}}\hspace{-6mm}
\subfigure{
\includegraphics[width=0.25\textwidth]{figures/Depth/depth_gcn_Pubmed0.png}}\\
\subfigure{
\includegraphics[width=0.25\textwidth]{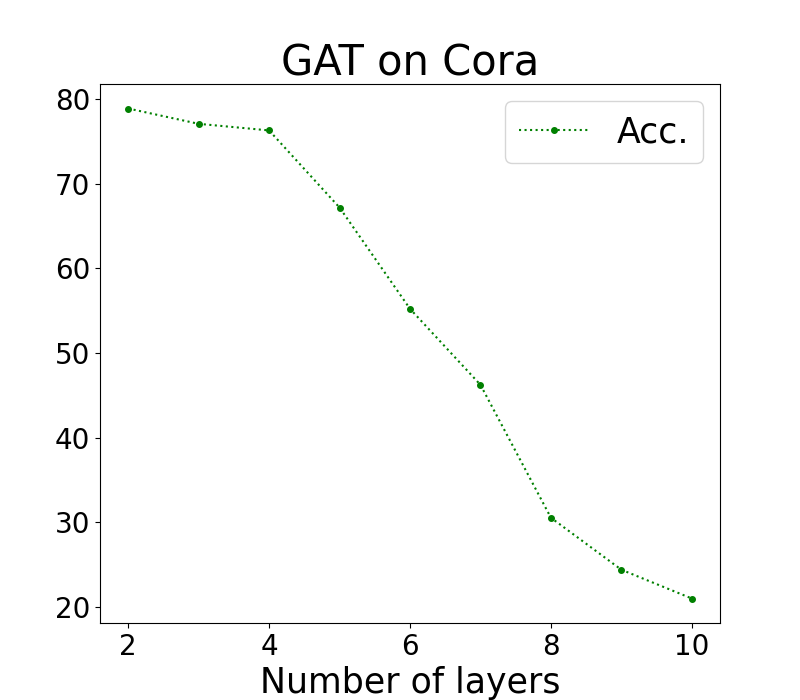}}\hspace{-6mm}
\subfigure{
\includegraphics[width=0.25\textwidth]{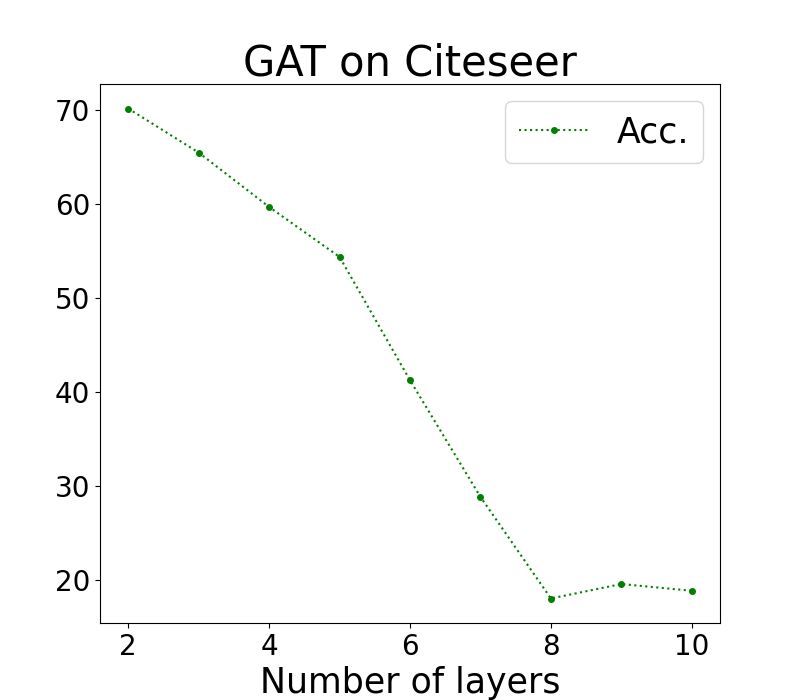}}\hspace{-6mm}
\subfigure{
\includegraphics[width=0.25\textwidth]{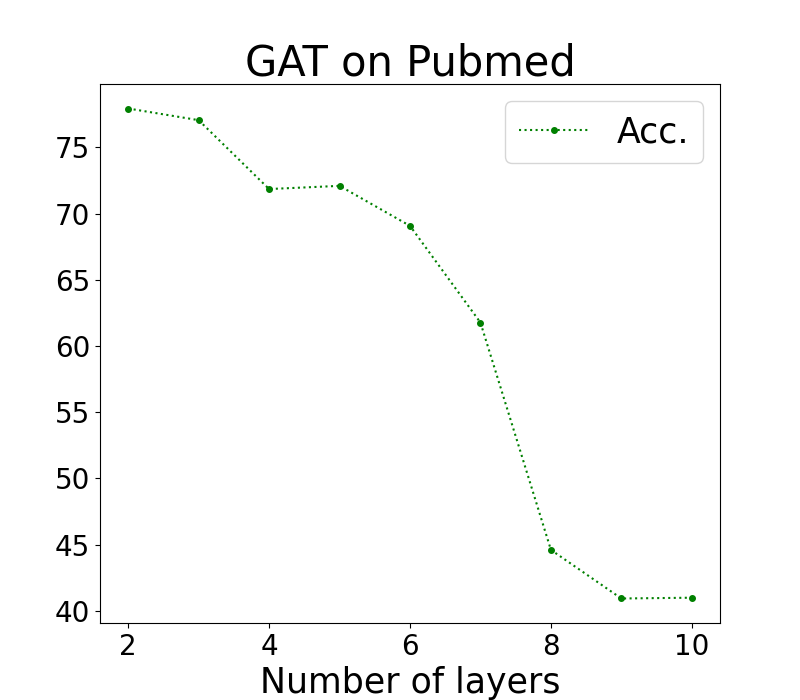}}
\\
\subfigure{
\includegraphics[width=0.25\textwidth]{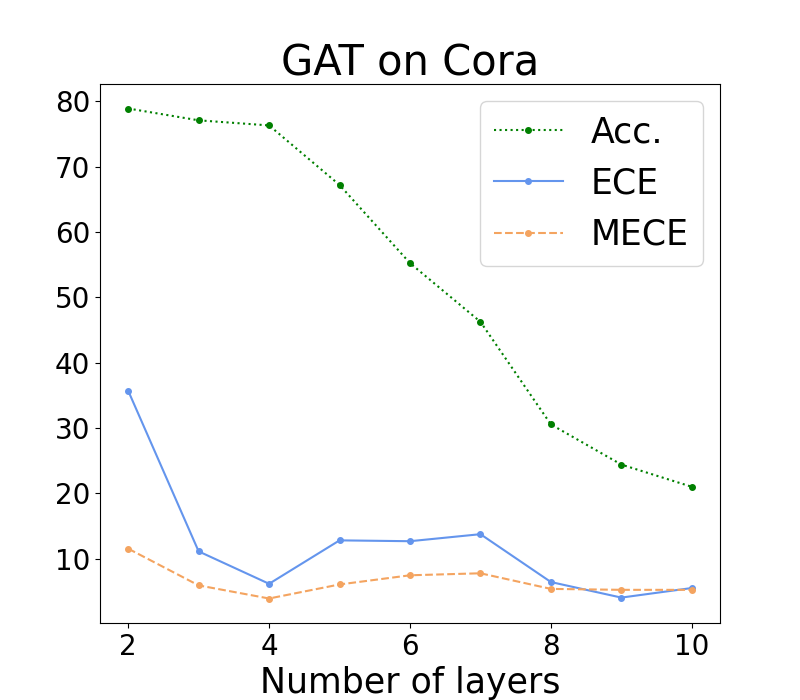}}\hspace{-6mm}
\subfigure{
\includegraphics[width=0.25\textwidth]{figures/Depth/depth_gat_Citeseer0.png}}\hspace{-6mm}
\subfigure{
\includegraphics[width=0.25\textwidth]{figures/Depth/depth_gat_Pubmed0.png}}\\
% \subfigure{
% \includegraphics[width=0.35\textwidth]{figures/Depth/depth_sgc_Cora1.png}}\hspace{-6mm}
% \subfigure{
% \includegraphics[width=0.35\textwidth]{figures/Depth/depth_sgc_Citeseer1.png}}\hspace{-6mm}
% \subfigure{
% \includegraphics[width=0.35\textwidth]{figures/Depth/depth_sgc_Pubmed1.png}}\\
% \vspace{-3mm}
% \subfigure{
% \includegraphics[width=0.35\textwidth]{figures/Depth/depth_sgc_Cora0.png}}\hspace{-6mm}
% \subfigure{
% \includegraphics[width=0.35\textwidth]{figures/Depth/depth_sgc_Citeseer0.png}}\hspace{-6mm}
% \subfigure{
% \includegraphics[width=0.35\textwidth]{figures/Depth/depth_sgc_Pubmed0.png}}\\
\caption{Influence of depth.}
\end{figure*}

\begin{figure*}[t]
\centering 
\subfigure{
\hspace{+3mm}\includegraphics[width=0.27\textwidth]{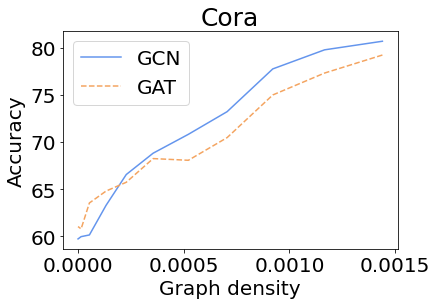}}\hspace{-6mm}
\subfigure{
\includegraphics[width=0.28\textwidth]{figures/Density/Cora_ECE.png}}\hspace{-3mm}
\subfigure{
\includegraphics[width=0.26\textwidth]{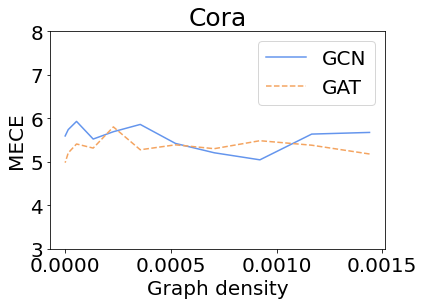}}\hspace{-5mm}\\
\subfigure{
\includegraphics[width=0.25\textwidth]{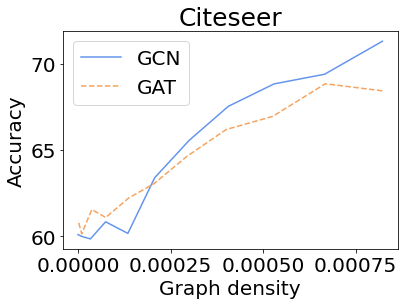}}
\subfigure{
\includegraphics[width=0.25\textwidth]{figures/Density/Citeseer_ECE.png}}
\subfigure{
\includegraphics[width=0.25\textwidth]{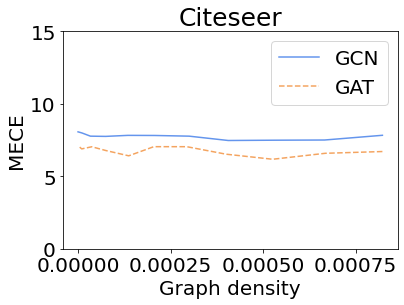}}\\
\subfigure{
\includegraphics[width=0.25\textwidth]{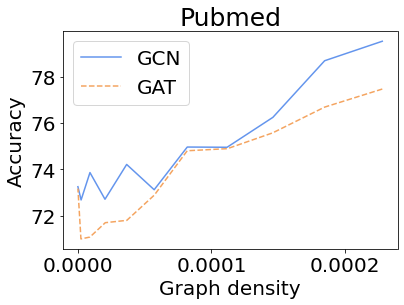}}
\subfigure{
\includegraphics[width=0.25\textwidth]{figures/Density/Pubmed_ECE.png}}
\subfigure{
\includegraphics[width=0.25\textwidth]{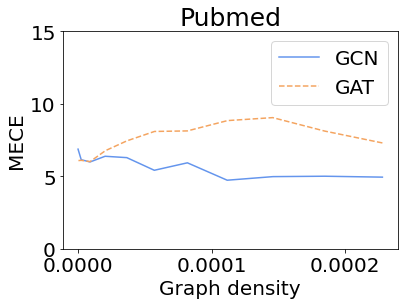}}
\caption{Influence of graph density.}
\end{figure*}

\end{document}